\pdfoutput=1

\documentclass[11pt]{article}

\usepackage[final]{EMNLP2023}

\usepackage{times}
\usepackage{latexsym}

\usepackage[T1]{fontenc}

\usepackage[utf8]{inputenc}

\usepackage{microtype}

\usepackage{inconsolata}

\usepackage{algorithm}
\usepackage{algpseudocode}
\usepackage{graphicx}
\usepackage{amsmath}
\usepackage{multirow}
\usepackage{listings}
\title{FAIR-RAG: Faithful Adaptive Iterative Refinement for Retrieval-Augmented Generation}

\author{Mohammad Aghajani Asl \\
  Sharif University of Technology \\
  \texttt{m.aghajani@physics.sharif.edu} \\
  \AND
  Majid Asgari-Bidhendi \\
  Iran University of Science and Technology \\
  Noor Avaran Jelvehaye Maanaei Najm Co., Ltd. \\
  \texttt{majid.asgari@gmail.com} \\
  \AND
  Behrooz Minaei-Bidgoli \\
  Iran University of Science and Technology \\
  \texttt{b\_minaei@iust.ac.ir} \\}

\begin{document}
\maketitle
\begin{abstract}
While Retrieval-Augmented Generation (RAG) mitigates hallucination and knowledge staleness in Large Language Models (LLMs), existing frameworks often falter on complex, multi-hop queries that require synthesizing information from disparate sources. Current advanced RAG methods, employing iterative or adaptive strategies, still lack a robust mechanism to systematically identify and fill evidence gaps, often propagating noise or failing to gather a comprehensive context. In this paper, we introduce FAIR-RAG, a novel agentic framework that transforms the standard RAG pipeline into a dynamic, evidence-driven reasoning process. At the core of FAIR-RAG is an Iterative Refinement Cycle governed by a novel module we term Structured Evidence Assessment (SEA). The SEA acts as an analytical gating mechanism: it deconstructs the initial query into a checklist of required findings and systematically audits the aggregated evidence to identify confirmed facts and, critically, explicit informational gaps. These identified gaps provide a precise, actionable signal to an Adaptive Query Refinement agent, which then generates new, targeted sub-queries to retrieve the missing information. This cycle repeats until the evidence is verified as sufficient, ensuring a comprehensive context for a final, strictly faithful generation step. We conduct extensive experiments on challenging multi-hop QA benchmarks, including HotpotQA, 2WikiMultiHopQA, and MusiQue. \textbf{Under a unified and controlled experimental setup,} FAIR-RAG significantly outperforms strong representative baselines. On HotpotQA, it achieves an F1-score of 0.453---an absolute improvement of 8.3 points over the strongest iterative baseline---\textbf{establishing a new state-of-the-art for this class of methods on these benchmarks.} Our work demonstrates that a structured, evidence-driven refinement process with explicit gap analysis is crucial for unlocking reliable and accurate reasoning in advanced RAG systems for complex, knowledge-intensive tasks.
\end{abstract}

\section{Introduction}

Large Language Models (LLMs) have demonstrated remarkable capabilities across a wide range of natural language processing tasks, including question-answering (QA)~\cite{brown2020languagemodelsfewshotlearners, chowdhery2022palmscalinglanguagemodeling}. However, their knowledge is inherently static, confined to the data they were trained on, which leads to factual inaccuracies and an inability to reason about events beyond their training cut-off date. Furthermore, LLMs are prone to ``hallucination,'' generating plausible yet factually incorrect information, which severely limits their reliability in knowledge-intensive applications~\cite{ji2023survey}. To mitigate these issues, Retrieval-Augmented Generation (RAG) has emerged as a prominent paradigm. By grounding the generation process on information retrieved from external knowledge bases, RAG systems aim to produce more accurate, timely, and verifiable responses~\cite{lewis2020retrieval}.

Despite its advantages, the standard ``retrieve-then-read'' RAG pipeline often falls short when faced with real-world user queries, which are frequently complex and cannot be answered through a single-shot retrieval step. For instance, a query such as \emph{``Which movie, directed by the same person who directed Inception, won an Oscar for Best Cinematography?''} requires multi-hop reasoning~\cite{yang2018hotpotqa}: first identifying the director of \emph{Inception} (Christopher Nolan) and then searching for his movies that have won the specified award. Standard RAG frameworks struggle with such multi-hop queries, as well as comparative or analytical questions that require synthesizing information from multiple sources. Moreover, they are not robust to suboptimal user query formulations and often fail to enforce that the generated answer remains strictly faithful to the retrieved evidence, thus perpetuating the risk of hallucination.

To address these limitations, several advanced RAG methodologies have been proposed. \textbf{Iterative approaches}, such as ITER-RETGEN~\cite{shao2023iter}, refine the retrieved information over multiple cycles by using the previously generated output as context for the next retrieval step. \textbf{Adaptive approaches}, like Adaptive-RAG~\cite{jeong2024adaptiveraglearningadaptretrievalaugmented}, aim for efficiency by dynamically selecting the retrieval strategy (e.g., no retrieval, single-step, or multi-step) based on an initial assessment of the query's complexity. Other frameworks like SELF-RAG~\cite{asai2023self} introduce self-reflection mechanisms, training the LLM to generate special tokens that control the retrieval and critique process on the fly. Nonetheless, a gap remains for a framework that synergistically combines iterative evidence refinement with adaptive query generation and an explicit, modular faithfulness check. Existing iterative methods can propagate noise by using entire generations as queries, while adaptive methods may not sufficiently refine the evidence needed for complex queries after the initial routing.

In this paper, we introduce \textbf{FAIR-RAG}, a novel framework for \textbf{F}aithful, \textbf{A}daptive, \textbf{I}terative \textbf{R}efinement in \textbf{R}etrieval-\textbf{A}ugmented \textbf{G}eneration. FAIR-RAG is designed to robustly handle complex queries by orchestrating a dynamic, multi-stage workflow. The framework begins with an \textbf{Adaptive Routing} module that analyzes query complexity to determine an optimal execution path, either by directly answering simple queries or by allocating the appropriate computational resources for complex ones. For non-trivial queries, FAIR-RAG initiates a cyclical process designed to progressively build and validate a context. At the core of this cycle is an \textbf{Iterative Refinement} loop where LLM agents intelligently decompose information needs, retrieve evidence, and filter out noise. Crucially, each cycle culminates in a \textbf{Structured Evidence Assessment (SEA)} module, which acts as an analytical gating mechanism. This module emulates a cognitive workflow by first deconstructing the user's query into a checklist of required findings. It then systematically synthesizes the retrieved evidence against this checklist, verifying what is confirmed and, most importantly, explicitly identifying any \textbf{``Remaining Gaps.''} These identified gaps provide a precise, actionable signal for the \textbf{Adaptive Query Refinement} module, which then generates new, targeted queries specifically designed to retrieve the missing information. This evidence-centric loop continues until sufficiency is achieved, ensuring that the final \textbf{Faithful Answer Generation} step is strictly grounded in a comprehensive and verified knowledge context, substantially enhancing trustworthiness and reducing hallucination.

Our main contributions are as follows:
\begin{itemize}
    \item We introduce a novel agentic RAG architecture centered on an \textbf{Iterative Refinement loop}. This evidence-centric cycle is governed by an analytical gating mechanism we term \textbf{Structured Evidence Assessment (SEA)}. By systematically deconstructing the query and identifying specific information gaps, the SEA module intelligently guides subsequent iterations. This process progressively builds and validates a comprehensive context, enabling the system to robustly handle complex, multi-faceted, and multi-hop queries where single-pass retrieval would fail.
    
    \item We design a sophisticated, two-stage query strategy. It begins with \textbf{semantic decomposition} to ensure all facets of the initial query are addressed, and more importantly, incorporates an \textbf{Adaptive Query Refinement} mechanism that analyzes evidence gaps to intelligently generate new queries, effectively reasoning about what information is still missing.
    
    \item We implement an integrated approach to resource optimization through \textbf{dynamic resource allocation}. Our framework employs an initial \textbf{Adaptive Routing} mechanism to bypass the RAG pipeline for simple queries and dynamically assigns LLMs of varying sizes to internal tasks based on their complexity, achieving a superior balance between response quality, latency, and computational cost.

    \item     We propose a robust, two-pronged approach to guarantee faithfulness. This includes: (1) a pre-generation \textbf{Structured Evidence Assessment (SEA)}, which performs a final analytical pass to verify that all required findings from the initial query deconstruction are fully supported by the aggregated evidence, and (2) a \textbf{constrained generation prompt} that enforces citation and prevents the model from introducing external knowledge. This combination ensures the final answer is both verifiable and trustworthy.
\end{itemize}

We conducted extensive experiments on a suite of four challenging open-domain QA benchmarks to evaluate the FAIR-RAG framework, encompassing complex multi-hop reasoning tasks (\textbf{HotpotQA}~\cite{yang2018hotpotqa}, \textbf{2WikiMultiHopQA}, \textbf{Musique}) and a large-scale single-hop factual dataset (\textbf{TriviaQA}). Our best-performing configuration, \textbf{FAIR-RAG 3 (Adaptive LLMs)}, was benchmarked against a wide range of strong baselines, including sequential (Standard RAG), conditional (Adaptive-RAG), and other state-of-the-art iterative methods (Iter-Retgen, Self-RAG).

The results clearly demonstrate the superiority of our approach. On the \textbf{HotpotQA} benchmark, our model sets a new state-of-the-art with an F1-score of \textbf{0.453}, surpassing the strongest iterative baseline, \textbf{Iter-Retgen (0.370)}, by a significant margin of \textbf{8.3 points}. This pattern of superior performance is consistent across other complex benchmarks: on \textbf{2WikiMultiHopQA}, FAIR-RAG achieves an F1 of \textbf{0.320}, outperforming the next best method, \textbf{Self-RAG (0.251)}, by \textbf{6.9 points}, and on \textbf{Musique}, it scores an F1 of \textbf{0.264}, which is \textbf{7.4 points} higher than \textbf{Iter-Retgen (0.190)}.

Notably, FAIR-RAG's architecture also excels on simpler factual queries, achieving a state-of-the-art F1 score of \textbf{0.731} on \textbf{TriviaQA}, showcasing its versatility. Our analysis further validates FAIR-RAG's core principles: performance consistently improves as the number of refinement iterations increases from one to three (e.g., F1 on HotpotQA improves from 0.398 for FAIR-RAG 1 to 0.447 for FAIR-RAG 3), confirming the value of the iterative evidence-gathering loop. Furthermore, the adaptive LLM allocation strategy provides an additional performance boost across all metrics. Finally, our framework consistently achieves the highest scores on the semantic metric \textbf{ACC\textsubscript{LLM}} (e.g., \textbf{0.847} on TriviaQA), confirming that its improvements reflect a deeper contextual understanding, not just lexical overlap. This robust performance validates the effectiveness of our iterative, evidence-driven framework in enhancing both the accuracy and faithfulness of LLM-based QA systems.

\section{Related Work}

The paradigm of Retrieval-Augmented Generation (RAG) has rapidly evolved from a straightforward retrieve-then-read pipeline to more sophisticated, dynamic frameworks. Our work, FAIR-RAG, builds upon and extends several key research threads in this domain.

\subsection{Standard Retrieval-Augmented Generation}

The foundational concept of RAG is to enhance the capabilities of LLMs by grounding them in external, non-parametric knowledge bases~\cite{lewis2020retrieval}. Early and influential RAG models typically employ a two-stage process: a retriever first fetches a set of relevant documents from a corpus based on the input query, and then a reader (or generator), which is often an LLM, synthesizes the final response conditioned on both the original query and the retrieved documents. This approach has proven effective in mitigating factual inconsistencies (hallucinations) and providing answers based on up-to-date information, thereby overcoming the static knowledge limitations of LLMs~\cite{Huang_2025,guu2020realm}. However, this single-shot retrieval mechanism is primarily designed for single-hop queries where the answer can be found within a small set of initially retrieved documents. Consequently, its performance degrades significantly on complex benchmarks like HotpotQA~\cite{yang2018hotpotqa}, which require multi-step reasoning or evidence aggregation from disparate sources.

\subsection{Iterative and Multi-Step RAG}

To address the shortcomings of standard RAG, a significant body of research has focused on iterative and multi-step approaches. These methods transform the single-shot process into a dynamic, multi-turn interaction. One prominent strategy involves decomposing a complex question into simpler sub-queries. While frameworks like \textbf{Self-Ask}~\cite{press2023measuringnarrowingcompositionalitygap} and \textbf{SuRe}~\cite{kim2024suresummarizingretrievalsusing} pioneer this decompositional approach, their strategies differ fundamentally from ours. \textbf{SuRe} generates a complete reasoning skeleton \emph{upfront}---a static plan that cannot adapt to the evidence retrieved in intermediate steps. \textbf{Self-Ask} generates sub-queries sequentially, but each new query is largely informed by the \emph{intermediate answer} to the previous one. \textbf{In contrast, FAIR-RAG's decomposition is neither static nor solely dependent on prior answers; it is a dynamic and context-aware process.} After each iteration, our framework uses the Structured Evidence Assessment (SEA) module to perform a holistic analysis of the \emph{entirety of the retrieved evidence corpus}. New sub-queries are then generated specifically to target the \textbf{explicitly identified informational gaps}. This gap-driven approach allows FAIR-RAG to adapt its evidence-gathering strategy in response to what has been found (and what is still missing), leading to a more robust and focused multi-step reasoning process than methods reliant on static plans or sequential, answer-driven prompting. 

Another popular approach interleaves reasoning steps with retrieval actions. This paradigm is exemplified by methods like ReAct~\cite{yao2022react} and IRCoT~\cite{trivedi2023interleavingretrievalchainofthoughtreasoning}, which guide an LLM to generate explicit reasoning traces (e.g., Chain-of-Thought~\cite{wei2022chain}), where each step can trigger a retrieval action to gather necessary information. The key distinction lies in the \textbf{scope and trigger} for retrieval. In methods like ReAct and IRCoT, retrieval is typically a \textbf{local, step-wise action} triggered by the immediate need of the next thought in a chain. In contrast, FAIR-RAG's retrieval is driven by a \textbf{holistic assessment} of the entire evidence pool against the overarching query requirements, allowing it to identify and address complex, non-sequential information gaps that step-wise reasoning might overlook.

More directly related to our work, frameworks like ITER-RETGEN~\cite{shao2023iter} propose a synergistic loop where the entire generated output from one iteration serves as the context to retrieve more relevant knowledge for the next. While powerful, using unstructured generation as a query can be suboptimal, as it may contain noise that misdirects the retriever. FAIR-RAG distinguishes itself by employing a more controlled mechanism: instead of using the entire previous output, it generates new, targeted sub-queries based on an explicit analysis of \emph{informational gaps}, leading to a more focused and efficient evidence-gathering process.

Other iterative frameworks have introduced forward-looking or corrective mechanisms to enhance retrieval precision. For instance, FLARE~\cite{jiang2023flare} employs a proactive strategy where the LLM anticipates future information needs during generation and triggers retrievals accordingly, effectively interleaving prediction and lookup steps. Similarly, Corrective RAG~\cite{yan2024corrective} incorporates a post-retrieval correction phase, using an evaluator to assess and refine retrieved documents based on their relevance and factual consistency. While these methods improve upon standard iterative loops by adding predictive or corrective elements, they often rely on heuristic triggers or post-hoc adjustments, which can still overlook systematic evidence gaps in highly complex queries. In contrast, FAIR-RAG's Structured Evidence Assessment (SEA) module provides a more principled, checklist-driven analysis that explicitly identifies and targets informational deficiencies, enabling a targeted and iterative refinement without dependence on generation-time predictions.

\subsection{Adaptive and Faithfulness-Aware RAG}

A third stream of research focuses on making RAG systems more adaptive and reliable. Adaptivity is often geared towards computational efficiency. For instance, \textbf{Adaptive-RAG}~\cite{jeong2024adaptiveraglearningadaptretrievalaugmented} introduces a classifier to pre-assess query complexity and route it to an appropriate strategy: no retrieval for simple questions, single-step retrieval for moderate ones, or a multi-step approach for complex queries. This routing is performed once at the beginning. In contrast, FAIR-RAG's adaptivity is dynamic and occurs \emph{within} the iterative process, as it continually adapts its query generation strategy based on the evolving set of retrieved evidence. 

Enhancing the faithfulness of the generated output is another critical concern. Standard RAG models do not explicitly guarantee that the generator will adhere to the retrieved context.~\cite{es2025ragasautomatedevaluationretrieval} To address this, SELF-RAG~\cite{asai2023self} fine-tunes an LLM to generate special ``reflection tokens'', enabling it to critique its own output for relevance and factual support in an inline fashion during generation. While effective, this approach has two limitations: its reliance on fine-tuning restricts applicability to off-the-shelf models, and its evaluation is inherently \textbf{tactical}, assessing evidence on a step-by-step basis as the answer is being composed. 

FAIR-RAG addresses the faithfulness challenge differently, through an explicit, modular, and more \textbf{strategic} Structured Evidence Assessment (SEA) module. Instead of an inline critique, SEA acts as a distinct analytical gating mechanism \emph{before} final answer generation. It first deconstructs the user's query into a checklist of required findings. It then performs a holistic audit of the \emph{entire} evidence corpus against this question-centric checklist to identify confirmed facts and explicit ``intelligence gaps.'' This distinction is crucial: whereas SELF-RAG prompts the model to ask, ``Is this next piece of evidence useful for my current generation step?'', SEA forces the model to first ask, ``Is the \emph{entirety} of my evidence sufficient to address all facets of the user's original query?'' The identified gaps from this strategic assessment provide a precise, actionable signal for the subsequent query refinement step, transforming the check from a passive validation into an active steering mechanism. Crucially, unlike fine-tuning-based methods, our modular SEA requires no model training, allowing for greater flexibility and easier integration with various off-the-shelf language models.

In summary, while existing works have made substantial advancements, FAIR-RAG provides a novel contribution by synergistically integrating three core principles into a cohesive framework: (1) a structured, gap-aware iterative refinement loop, (2) context-aware, adaptive sub-query generation, and (3) an explicit, modular faithfulness assessment that requires no model fine-tuning.

\section{Methodology}

\begin{figure*}[t]
\centering
\includegraphics[width=\textwidth]{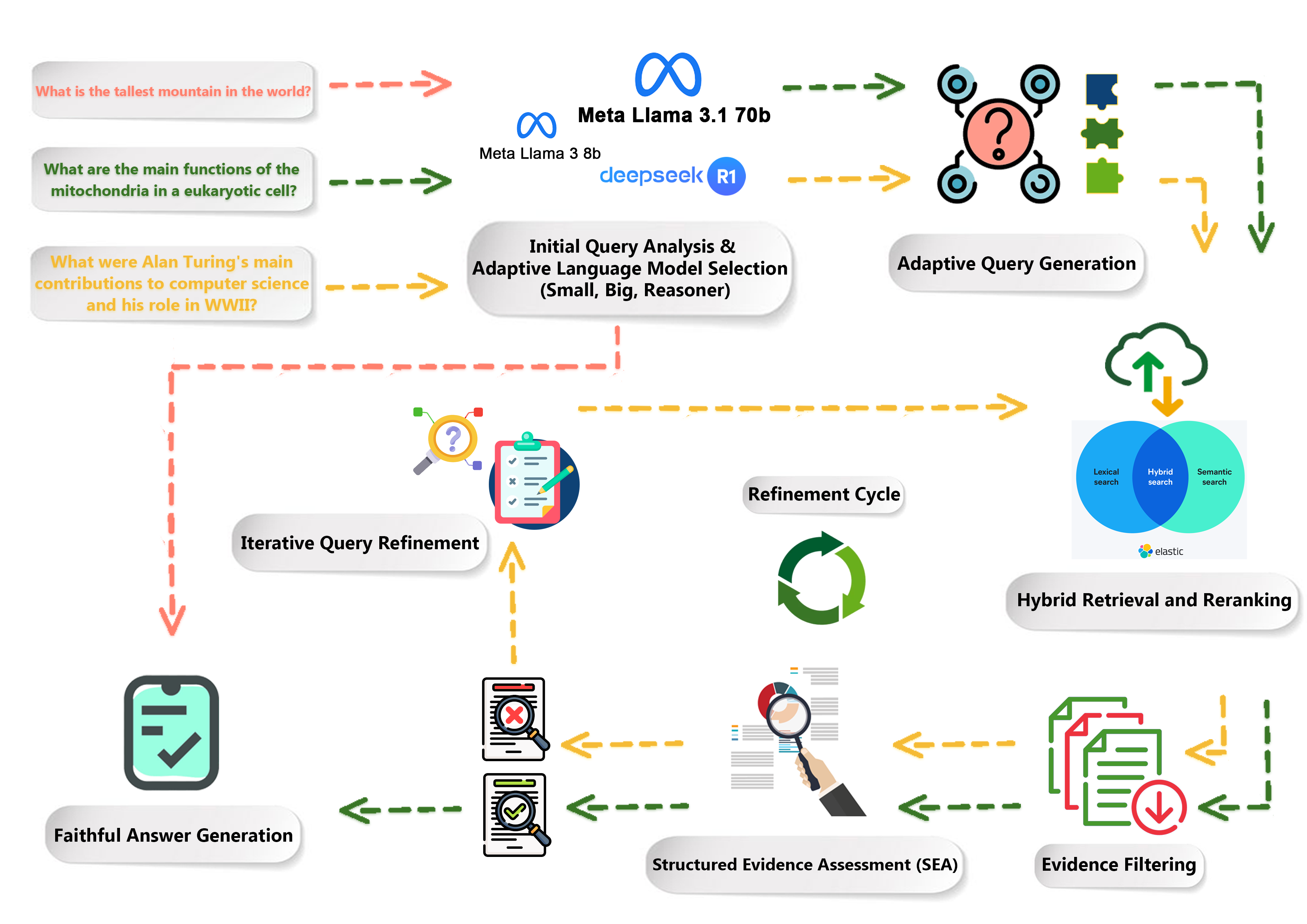}
\caption{Schematic overview of the FAIR-RAG architecture. The process starts with initial query analysis and adaptive language model selection (e.g., small, large, or reasoning LLM). For complex queries, it proceeds to query decomposition, followed by an iterative refinement cycle involving hybrid retrieval and reranking, evidence filtering, and Structured Evidence Assessment (SEA). The loop iterates until evidence sufficiency is confirmed, culminating in faithful answer generation grounded in the aggregated evidence.}
\label{fig:architecture}
\end{figure*}

The FAIR-RAG framework is designed as a multi-stage, iterative process that dynamically adapts its strategy to the complexity of a user's query. Our architecture transforms the standard, static RAG pipeline into an intelligent, evidence-driven workflow that progressively builds context to answer complex questions. The entire process, illustrated in Figure~\ref{fig:architecture}, can be divided into four main phases: (1) Initial Query Analysis and Adaptive Routing, (2) The Iterative Retrieval and Refinement Cycle, (3) Faithful Answer Generation, and (4) Dynamic Resource Allocation. The entire inference procedure is detailed step-by-step in Algorithm~\ref{alg:fairrag}. The process is initiated by the \textbf{Adaptive Routing} agent ($A_{\text{router}}$), which classifies the query and selects the most appropriate generator LLM ($G_{\text{selected}}$) from the predefined set of available models ($\mathcal{G}$).

\begin{algorithm*}[t]
\caption{FAIR-RAG Inference (Prompts for agents are detailed in Appendix B.)}
\label{alg:fairrag}
\small
\begin{algorithmic}[1]
\Require Set of Generator LLMs $\mathcal{G} = \{G_{\text{small}}, G_{\text{large}}, G_{\text{reasoning}}\}$, Hybrid Retriever $H_{\text{retriever}}$, Document Corpus $\mathcal{C} = \{d_1, \ldots, d_N\}$, Set of LLM Agents $\mathcal{A} = \{A_{\text{router}}, A_{\text{decompose}}, A_{\text{filter}}, A_{\text{suff}}, A_{\text{refine}}\}$
\State \textbf{Input:} User query $x$
\State \textbf{Output:} Final, evidence-grounded answer $y_{\text{final}}$
\State $\text{query\_type}, G_{\text{selected}} \leftarrow A_{\text{router}}.\text{classify}(x)$ \hfill $\triangleright$ Route
\If{$\text{query\_type} == \text{OBVIOUS}$}
    \State $y_{\text{final}} \leftarrow G_{\text{selected}}.\text{generate}(x)$ \hfill $\triangleright$ Generate directly from parametric knowledge
    \State \Return $y_{\text{final}}$
\EndIf
\State $E_{\text{agg}} \leftarrow \emptyset$ \hfill // Aggregated evidence
\State $Q_{\text{previous}} \leftarrow \{x\}$
\For{$i \leftarrow 1$ \textbf{to} $3$} \hfill // Max 3 iterations
    \If{$i == 1$}
        \State $Q_{\text{sub}} \leftarrow A_{\text{decompose}}.\text{generate}(x)$ \hfill $\triangleright$ Decompose initial query
    \Else
        \State $Q_{\text{sub}} \leftarrow A_{\text{refine}}.\text{generate}(x, Q_{\text{previous}}, \text{analysis\_summary})$ \hfill $\triangleright$ Refine query to fill gaps
    \EndIf
    \State $Q_{\text{previous}} \leftarrow Q_{\text{previous}} \cup Q_{\text{sub}}$
    \State $D_{\text{candidate}} \leftarrow H_{\text{retriever}}.\text{retrieve\_and\_rerank}(Q_{\text{sub}}, \mathcal{C})$ \hfill $\triangleright$ Retrieve
    \State $E_{\text{filtered}} \leftarrow A_{\text{filter}}.\text{filter}(D_{\text{candidate}}, x)$ \hfill $\triangleright$ Filter irrelevant evidence
    \State $E_{\text{agg}} \leftarrow E_{\text{agg}} \cup E_{\text{filtered}}$
    \State $\text{analysis\_summary}, \text{is\_sufficient} \leftarrow A_{\text{suff}}.\text{check}(E_{\text{agg}}, x)$ \hfill $\triangleright$ Assess sufficiency with SEA
    \If{$\text{is\_sufficient} == \text{Yes}$}
        \State \textbf{break} \hfill // Exit loop if evidence is sufficient
    \EndIf
\EndFor
\State $y_{\text{final}} \leftarrow G_{\text{selected}}.\text{generate}(x, E_{\text{agg}}, \text{faith\_constraints})$ \hfill $\triangleright$ Generate answer
\State \Return $y_{\text{final}}$
\end{algorithmic}
\end{algorithm*}

\subsection{Overall Architecture}

The overall architecture of FAIR-RAG, illustrated in Figure~\ref{fig:architecture}, employs a dynamic, multi-step process to handle user queries. An initial routing step assesses query complexity. Simple queries are answered directly, while complex ones trigger an iterative refinement loop. This core loop consists of adaptive query generation, hybrid retrieval, filtering, and a Structured Evidence Assessment (SEA). The SEA module determines if the collected evidence is sufficient. If not, the loop repeats with refined queries targeting information gaps. Once sufficiency is met, a final, evidence-grounded answer is generated. The entire inference procedure is detailed step-by-step in Algorithm~\ref{alg:fairrag}. All LLM agents in our pipeline are guided by meticulously engineered prompts. The full details and examples for each prompt are provided in Appendix B. 

\subsection{Initial Query Analysis and Adaptive Routing}

The first stage of our framework is a lightweight yet crucial analysis of the input query. An LLM agent classifies the query into one of four categories to determine the subsequent workflow and resource allocation:

\begin{itemize}
    \item \textbf{OBVIOUS:} For queries whose answers are likely stable and contained within the LLM's parametric knowledge (e.g., ``What is the capital of France?''). These queries are routed directly to a large LLM for generation, bypassing the entire RAG pipeline for maximum efficiency.
    \item \textbf{SMALL:} Simple factual queries that require retrieval but minimal reasoning. A smaller, more efficient LLM is designated for the final generation step.
    \item \textbf{LARGE:} Queries that require retrieval and synthesis of information from multiple sources. A larger, more capable LLM is selected for generation.
    \item \textbf{REASONING:} Complex queries that demand multi-hop reasoning, comparison, or deep analysis. A state-of-the-art LLM with strong reasoning capabilities is allocated for the final answer.
\end{itemize}

This initial routing mechanism serves a dual purpose: it acts as an efficiency-enhancing shortcut for simple queries and implements our \textbf{Adaptive Model Selection} strategy by pre-allocating the most cost-effective generator model for the final step. While this module is a key feature for practical deployment, it was systematically controlled during our benchmark evaluations to ensure a fair comparison, as detailed in Section 4.4. 

\subsection{The Iterative Retrieval and Refinement Cycle}

This cycle is the core of FAIR-RAG's ability to handle complex information needs. It is designed to run for a maximum of three iterations to ensure a balance between comprehensiveness and latency. Each iteration consists of the following steps:

\subsubsection{Adaptive Query Generation}

The initial user query, if deemed complex, is first subjected to semantic decomposition. An LLM agent breaks down the multifaceted query into a set of up to four distinct, keyword-rich, and semantically independent sub-queries. For example, the query \textbf{``What were Alan Turing's main contributions to computer science and his role in World War II?''} is decomposed into targeted sub-queries like \textbf{``Alan Turing's contributions to theoretical computer science''} and \textbf{``Alan Turing's role in breaking the Enigma code.''} This ensures that the retrieval process covers all conceptual facets of the original question. 

\subsubsection{Hybrid Retrieval and Reranking}

For each sub-query, we employ a hybrid retrieval strategy to maximize recall. We perform both dense vector search (capturing semantic similarity)~\cite{karpukhin2020dense} and traditional keyword-based sparse search (capturing exact matches)~\cite{robertson2009probabilistic}. The results from both methods are aggregated, and the documents are re-ranked using the \textbf{Reciprocal Rank Fusion (RRF)} algorithm~\cite{cormack2009reciprocal}. RRF effectively combines the rankings from both retrieval methods without requiring hyperparameter tuning, producing a single, robustly ranked list of the top-5 most relevant documents as candidate evidence.

\subsubsection{Evidence Filtering}

The candidate evidence is then passed to a filtering module. An LLM agent evaluates each document's utility with respect to the \emph{original} user query. Documents that are irrelevant, off-topic, or only tangentially related are discarded~\cite{liu2023lostmiddlelanguagemodels}. This step is critical for increasing the signal-to-noise ratio of the context provided to the final generator, preventing the model from being distracted by noisy or unhelpful information. 

\subsubsection{Structured Evidence Assessment (SEA): The Strategic Intelligence Analyst}

The strategic core of our iterative refinement loop is the Structured Evidence Assessment (SEA) module. Its primary function is not merely to verify evidence, but to perform a granular \textbf{gap analysis} that generates an explicit, actionable signal for the next iteration. For this critical task, we deliberately chose a \textbf{checklist-based methodology} over alternatives like abstractive summarization or direct question-answering. Abstractive summarization, by design, often conceals the very gaps we need to identify by creating a fluent narrative, while direct QA yields binary outcomes that fail to pinpoint the precise location of missing information in a multi-fact query.

Our approach operationalizes this checklist via an LLM agent prompted to act as a \textbf{Strategic Intelligence Analyst}. Through carefully engineered prompts---defining its role and providing few-shot examples (see Appendix B)---the agent first \textbf{deconstructs} the user's query into a checklist of discrete, required informational components or ``findings.'' It then systematically \textbf{audits} the collected evidence against this checklist, confirming which findings are supported and identifying which remain as explicit \textbf{``intelligence gaps.''} The evidence is deemed sufficient only if all required findings are confirmed.

The `unchecked' items on this list constitute a \textbf{direct, interpretable, and actionable signal} for the subsequent Query Refinement module. This structured, question-centric process provides a more \textbf{controllable, reliable, and transparent mechanism} for gap analysis than less constrained methods, ensuring a rigorous evaluation that prevents the system from being misled by large volumes of tangentially related information.

\subsubsection{Iterative Query Refinement}

If the Structured Evidence Assessment (SEA) results that the evidence is insufficient to correctly answer the question, the system activates the query refinement module, which is designed to be a highly targeted intervention. An LLM agent uses the ``Remaining Gaps'' and ``Confirmed Findings'' from the analyst's summary (generated in the previous step) as its primary input. Its goal is to generate new, laser-focused queries that are engineered to find \textbf{only} the missing pieces of information. By leveraging the confirmed findings, the agent makes the new queries more precise and avoids repeating previous searches. For instance, if a query is \textbf{``In what city was the lead scientist who broke the Enigma code buried?''} and the analyst's summary confirms \textbf{``Alan Turing was the lead scientist''} but identifies \textbf{``Alan Turing's burial place''} as a gap, the refinement module will generate new, highly targeted queries like \textbf{``Alan Turing burial place''} or \textbf{``city where Alan Turing is buried.''} These new queries, now contextualized and specific, re-enter the hybrid retrieval cycle (Step 3.3.2). 

\subsection{Faithful Answer Generation}

Once the Structured Evidence Assessment (SEA) returns ``Yes,'' the curated and validated evidence set is passed to the generator LLM selected in the initial routing stage (Step 3.2). The generation process is governed by a meticulously engineered prompt that enforces a \textbf{strict, evidence-only generation protocol}. This constrained prompt is designed to maximize faithfulness and minimize the risk of hallucination by instructing the LLM to:

\begin{itemize}
    \item Base its answer \textbf{exclusively} on the provided evidence.
    \item Avoid introducing any external information, parametric knowledge, or opinions.
    \item Cite every claim by embedding reference tokens (e.g., [1], [2]) that link to the source documents.
    \item If the evidence is ultimately insufficient, state this directly without attempting to speculate.
\end{itemize}

This highly constrained generation process is fundamental to mitigating hallucination and ensures that the final output is not only accurate but also transparent and fully traceable to its sources. 

\subsection{Dynamic Resource Allocation and Prompt Engineering}

A key aspect of FAIR-RAG's design is the efficient use of computational resources. We dynamically allocate LLMs of different sizes (e.g., SMALL, LARGE) for the various internal tasks based on their complexity. For instance, simpler tasks like the \textbf{Initial Query Analysis and Adaptive Routing} are handled by a smaller model, while the highly nuanced task of \textbf{query refinement} is assigned to a more capable model to ensure maximum precision. This dynamic allocation optimizes the trade-off between performance, cost, and latency.

Furthermore, all interactions with LLM agents are managed through a \textbf{structured prompt engineering methodology} designed to ensure reliability and predictability [see Appendix B for details]. Our prompts are consistently engineered to include several key components that guide the model's behavior:

\begin{itemize}
    \item \textbf{Role and Context Definition:} Each prompt begins by establishing a clear role for the agent and the context of its task (e.g., \emph{``You are a Strategic Intelligence Analyst\ldots''}).
    \item \textbf{Task Specification:} The primary goal or intent of the task is explicitly stated (e.g., \emph{``Your mission is to determine if the provided evidence is sufficient\ldots''}).
    \item \textbf{Guided Reasoning (Scaffolding):} We provide clear, step-by-step instructions, analytical guidelines, or few-shot examples to structure the model's reasoning process and ensure consistency.
    \item \textbf{Behavioral Constraints:} Explicit rules are laid out that the model must follow, governing its process and output (e.g., \emph{``You MUST follow this thinking process and output format exactly''}).
    \item \textbf{Output Formatting:} The exact format for the response is strictly defined to ensure a machine-parseable and predictable output (e.g., \emph{``A single word: `Yes' or `No'''}).
\end{itemize}

This robust, component-based prompting strategy ensures that the LLM agents perform their designated roles reliably, contributing to the overall stability and performance of the framework~\cite{liu2021pretrainpromptpredictsystematic}.

\section{Experimental Setup}

To rigorously evaluate the performance of FAIR-RAG, we conduct a series of experiments on challenging question-answering benchmarks. Our experimental protocol is built upon the standardized and open-source \textbf{FlashRAG} toolkit, ensuring a fair and reproducible comparison against existing methods.

\subsection{Datasets}

We selected a carefully curated suite of four benchmark datasets to assess the various capabilities of our framework, with a particular focus on complex, multi-hop reasoning where standard RAG systems often underperform.

\begin{itemize}
    \item \textbf{Multi-hop QA:} We use \textbf{HotpotQA}, \textbf{2WikiMultihopQA}, and \textbf{MusiQue}. These datasets are specifically designed to require reasoning and synthesizing information across multiple documents to arrive at an answer, making them ideal for evaluating our iterative refinement and adaptive query generation modules.
    \item \textbf{Open-Domain QA:} We also use \textbf{TriviaQA}, a popular dataset for open-domain question answering, to ensure our model maintains strong performance on simpler, fact-based queries.
\end{itemize}

For all experiments, we follow the standard practice of using the official test split where available; otherwise, we report results on the development split. Consistent with recent studies on computationally intensive RAG models, and to manage the substantial API and computational costs associated with iterative inference frameworks like ours, all evaluations are conducted on a randomly selected subset of 1000 samples from each dataset. \textbf{This sample size was deliberately chosen to strike a critical balance between statistical robustness and experimental feasibility.} It is large enough to ensure stable and meaningful performance comparisons while enabling the comprehensive suite of ablation studies and baseline comparisons presented in this work, which would be financially and logistically intractable on the full datasets.

\begin{table}[t]
\centering
\small
\begin{tabular}{@{}lccc@{}}
\hline
\textbf{Dataset} & \textbf{Task Type} & \textbf{Source} & \textbf{Samples} \\
\hline
HotpotQA & multi-hop QA & wiki & 1000 \\
2WikiMultiHopQA & multi-hop QA & wiki & 1000 \\
MusiQue & multi-hop QA & wiki & 1000 \\
TriviaQA & Open-Domain & wiki/web & 1000 \\
\hline
\end{tabular}
\caption{Summary of Datasets}
\label{tab:datasets}
\end{table}

\subsection{Baselines}

We compare FAIR-RAG against a comprehensive set of representative RAG baselines implemented within the FlashRAG framework. These baselines cover different architectural paradigms:

\begin{itemize}
    \item \textbf{Standard RAG}~\cite{lewis2020retrieval}: A conventional retrieve-then-read pipeline serving as a fundamental baseline.
    
    \item \textbf{Iterative Methods:} We include \textbf{Iter-Retgen (ITRG)}~\cite{shao2023iter}, which uses the previous generation's output to retrieve new documents, and \textbf{IRCoT}~\cite{trivedi2023interleavingretrievalchainofthoughtreasoning}, which integrates retrieval within a Chain-of-Thought process.
    
    \item \textbf{Reasoning-based Methods:} We compare against \textbf{ReAct}~\cite{yao2022react}, a popular agent-based framework that interleaves reasoning and action steps to solve problems.
    
    \item \textbf{Faithfulness-focused Methods:} \textbf{Self-RAG}~\cite{asai2023self} is included as a strong baseline that incorporates explicit reflection and self-critique steps to improve faithfulness.
    
    \item \textbf{Branching \& Conditional Methods:} We include \textbf{SuRe}~\cite{kim2024suresummarizingretrievalsusing}, which generates and ranks multiple candidate answers, and \textbf{Adaptive-RAG}~\cite{jeong2024adaptiveraglearningadaptretrievalaugmented}, which uses a classifier to conditionally route queries through different execution paths.
\end{itemize}

\subsection{Evaluation Metrics}

To provide a comprehensive assessment of our system's performance, we employ a suite of metrics that capture both lexical accuracy and semantic correctness.

\subsubsection{Lexical-Based Metrics}

Following standard practice for question-answering tasks, we first report two traditional, token-based metrics:

\begin{itemize}
    \item \textbf{Exact Match (EM):} This strict metric measures the percentage of predictions that match one of the ground-truth answers exactly, character for character.
    \item \textbf{F1 Score:} A more lenient metric that computes the harmonic mean of precision and recall at the token level. It accounts for partial overlaps and is less sensitive to minor phrasing differences than EM.
\end{itemize}

\subsubsection{Automated Evaluation using LLM-as-Judge}

Lexical metrics like EM and F1 are often insufficient for evaluating generative models, as they unfairly penalize semantically correct answers that are phrased differently or contain additional, relevant context not present in the ground truth. To overcome this limitation, we employ LLM-as-Judge methodologies for two distinct, nuanced evaluation purposes, strategically selecting the judge model based on task complexity.

\begin{itemize}
    \item \textbf{End-to-End Semantic Correctness (ACC\textsubscript{LLM}):} For the large-scale evaluation of our main results (Table 2), we introduce \textbf{LLM-as-Judge Accuracy (ACC\textsubscript{LLM})}. This metric requires a scalable and consistent binary judgment on whether a final prediction is semantically equivalent to any ground-truth answer. For this task, we use the highly capable \textbf{``Meta-Llama-3-8B-Instruct''} model as the judge.~\cite{meta2024llama3} The specific prompt, designed for efficient ``Yes'' or ``No'' classification, is detailed in Appendix C.1~\cite{vicuna2023,Zheng2023Judging}.
    
    \item \textbf{Component-Level Quality Score:} Given the nuanced, generative nature of intermediate outputs in our ablation study (e.g., Query Decomposition), a simple binary metric is insufficient. Therefore, we employ an LLM-as-Judge methodology~\cite{Zheng2023Judging}, building on established frameworks like G-Eval~\cite{liu2023gevalnlgevaluationusing} to ensure a scalable and consistent assessment. For this role, we selected Llama-4-Maverick-17B-128E-Instruct-FP8.~\cite{meta2025llama4} This highly capable model was specifically chosen not merely for its general performance, but for its demonstrated aptitude in complex reasoning and nuanced instruction following. These capabilities are critical for accurately assessing the quality of our internal components, where evaluation criteria are intricate and context-dependent. The reliability of this model as a proxy for human judgment in our specific tasks is not an unsubstantiated claim; as we will demonstrate in our validation study (Section 5.2.1), our LLM-as-Judge's ratings show a strong correlation with those of human experts, confirming its suitability for this evaluation. To ensure high-quality, structured feedback and mitigate potential biases, we designed custom prompts for each component. These prompts provide the judge with clear task definitions, explicit scoring criteria on a 1-to-5 scale, and illustrative examples. This rigorous, prompt-driven approach, detailed in Appendix C.2, ensures consistent and interpretable scores for a credible and multi-faceted analysis of our pipeline's internal mechanics.
\end{itemize}

\subsubsection{Reliability of LLM-as-Judge Evaluations}

To establish the credibility of our dual LLM-as-Judge framework, we conducted two separate human verification studies, one for each evaluation type.

\begin{itemize}
    \item \textbf{Verification of Binary Semantic Correctness (ACC\textsubscript{LLM}):} The reliability of the Llama-3-8B-Instruct judge, used for the ACC\textsubscript{LLM} metric, was rigorously validated. A random subset of \textbf{300} question-answer pairs was sampled, stratified across all datasets. A human expert, \textbf{blinded to the LLM's original decision to prevent bias}, annotated these pairs for semantic correctness. The human judgments showed a \textbf{strong degree of concordance} with the LLM-as-Judge's binary ``Yes/No'' outputs, aligning in \textbf{90\%} of the cases. This level of agreement is well within the range of typical human inter-annotator agreement for such a nuanced task. Therefore, we conclude that the LLM-as-Judge serves as a reliable and scalable proxy for human evaluation in our experiments.
    
    \item \textbf{Verification of Component-Level Quality Scores:} For the more nuanced 1-to-5 scale scoring performed by the powerful Llama-4-Maverick model in our ablation study, a similar validation was conducted. A separate random subset of 100 generated outputs from the component-level tasks (e.g., query refinement) was evaluated by a human expert, again blinded to the LLM's score. The evaluation demonstrated a \textbf{95\%} agreement between human and LLM judgments across these different tasks. This confirms the judge's capability for providing consistent, human-aligned quality assessments.
\end{itemize}

This dual validation confirms that our LLM-as-Judge methodologies, for both binary correctness and fine-grained quality scoring, serve as robust and reliable proxies for human evaluation, enabling a credible and scalable assessment of our framework's performance.

\subsection{Implementation Details}

To ensure a controlled and fair comparison, all experiments adhere to the global settings defined by the FlashRAG framework, with specific adaptations for our models.

\paragraph{Methodological Alignment for Fair Comparison:}

A cornerstone of our evaluation is the principle of fair comparison, ensuring that reported performance gains are attributable to our core architectural innovations (i.e., the iterative refinement cycle) rather than peripheral components. Our full proposed architecture includes features designed for optimal real-world efficiency, such as a hybrid retriever (Section 3.3.2) and an ``OBVIOUS'' query shortcut (Section 3.2). However, as these features are not standard in the baseline methods we compare against, they could introduce an unfair advantage.

Therefore, to create a level playing field, \textbf{these two capabilities were systematically disabled during all benchmark experiments for all methods, including our FAIR-RAG variants.} Specifically:

\begin{itemize}
    \item All models used a standardized \textbf{dense-only retriever}.
    \item The ``OBVIOUS'' query routing was deactivated, forcing \textbf{every query to pass through the full RAG pipeline.}
\end{itemize}

This rigorous alignment ensures that the observed performance differences are a direct result of the models' reasoning and evidence-handling capabilities.

\paragraph{Component Configuration:}

\begin{itemize}
    \item \textbf{Retriever:} For all methods, we use \textbf{e5-base-v2}~\cite{wang2024textembeddingsweaklysupervisedcontrastive} as the sole dense retriever, configured to fetch the top 5 documents per query from the standard DPR Wikipedia (Dec. 2018)~\cite{karpukhin2020dense} corpus. The index is built using Faiss (Flat type)~\cite{johnson2019billion} to ensure accuracy.
    
    \item \textbf{Generator Models:} To benchmark against a powerful and widely accessible model, we standardize the generator for all baseline methods to be ``Llama-3-8B-Instruct'', accessed via API. A key exception is Self-RAG, for which we use its officially released, fine-tuned selfrag-llama-7b model to respect its original design.
    
    \item \textbf{FAIR-RAG Configuration:} Our experimental setup for FAIR-RAG is designed to ensure a rigorous and fair comparison. We evaluate two primary configurations of our framework:
    \begin{itemize}
        \item \textbf{Uniform Model Configuration:} To isolate the architectural benefits of our iterative approach, the FAIR-RAG 1-4 variants exclusively use Llama-3-8B-Instruct for all internal tasks and for the final answer generation. This ensures a direct and fair comparison against all baseline methods, which are also benchmarked using the same Llama-3-8B-Instruct model. In this configuration, the adaptive LLM selection (SMALL, LARGE, REASONING roles) is intentionally disabled, with all roles defaulting to the single model.
        
        \item \textbf{Adaptive LLM Configuration:} To demonstrate the full potential of our framework, we also report results for FAIR-RAG (Adaptive LLMs). This configuration employs a dynamic, multi-agent allocation strategy to optimize the trade-off between performance and cost:
        \begin{itemize}
            \item For less complex internal tasks, such as query decomposition and Structured Evidence Assessment (SEA), we utilize Llama-3-8B-Instruct.
            \item For more cognitively demanding tasks, including evidence filtering, query refinement, and faithful answer generation, we leverage the more powerful Llama-3.1-70B-Instruct.~\cite{meta2024llama3}
            \item For tasks requiring deep reasoning, the system routes to a specialized DeepSeek-R1 model.~\cite{deepseek2025r1}
        \end{itemize}
    \end{itemize}
\end{itemize}

Unless otherwise specified, all other hyperparameters adhere to the default settings of the underlying framework to maintain consistency across experiments.

\section{Results}

This section presents a comprehensive evaluation of FAIR-RAG. We first report the main end-to-end results against a suite of strong baseline methods (Section~\ref{sec:main-results}). We then provide a deeper analysis of the framework's internal mechanics, including a component-wise ablation study (Section~\ref{sec:ablation}) and an examination of the impact of iterative refinement on answer quality and cost (Section~\ref{sec:iterative-impact}).

\subsection{Main Results}
\label{sec:main-results}

Table~\ref{tab:main-results} presents the main performance comparison of FAIR-RAG and baseline methods across four diverse question-answering benchmarks. Our framework demonstrates state-of-the-art performance, particularly on datasets that require complex, multi-step reasoning. We report four key metrics: \textbf{Exact Match (EM)} and \textbf{F1-Score} for lexical accuracy, a script-based \textbf{Accuracy (ACC)} which measures the presence of the ground-truth answer string in the generation, and our primary semantic metric, \textbf{LLM-as-Judge Accuracy (ACC\textsubscript{LLM})}. While EM, F1, and ACC are token-based, \textbf{ACC\textsubscript{LLM}} evaluates semantic equivalence, offering a more robust assessment of generative answers.

To provide a fine-grained analysis of our framework's iterative capabilities, we evaluate several configurations of FAIR-RAG in Table~\ref{tab:main-results}. The FAIR-RAG 1 to 4 variants correspond to the system's performance with the maximum number of iterations capped at 1 to 4, respectively. To ensure a fair comparison against the baselines and to isolate the impact of the iterative refinement cycle itself, these variants utilize a single, consistent generator model (Llama-3-8B-Instruct) across all stages. Conversely, the (Adaptive LLMs) variants represent our full, optimized framework, employing the dynamic allocation of different LLM agents (e.g., small, large, reasoner) based on task complexity, as detailed in our methodology.

\begin{table*}[t]
\centering
\scriptsize
\setlength{\tabcolsep}{3pt}
\begin{tabular}{llcccccccccccccccc}
\hline
\multirow{2}{*}{\textbf{Type}} & \multirow{2}{*}{\textbf{Method}} & \multicolumn{4}{c}{\textbf{HotpotQA}} & \multicolumn{4}{c}{\textbf{2WikiMultiHopQA}} & \multicolumn{4}{c}{\textbf{Musique}} & \multicolumn{4}{c}{\textbf{TriviaQA}} \\
\cline{3-18}
& & \textbf{EM} & \textbf{F1} & \textbf{ACC} & \textbf{ACC\textsubscript{LLM}} & \textbf{EM} & \textbf{F1} & \textbf{ACC} & \textbf{ACC\textsubscript{LLM}} & \textbf{EM} & \textbf{F1} & \textbf{ACC} & \textbf{ACC\textsubscript{LLM}} & \textbf{EM} & \textbf{F1} & \textbf{ACC} & \textbf{ACC\textsubscript{LLM}} \\
\hline
Sequential & Standard RAG & .238 & .342 & .328 & .598 & .086 & .180 & .282 & .369 & .074 & .149 & .137 & .380 & .570 & .667 & .675 & .795 \\
Branching & Sure & .232 & .346 & .293 & .646 & .134 & .198 & .167 & .408 & .101 & .169 & .114 & .397 & .566 & .675 & .641 & .799 \\
Conditional & Adaptive-RAG & .144 & .239 & .368 & .628 & .041 & .133 & .340 & .429 & .038 & .095 & .160 & .366 & .480 & .583 & .670 & .789 \\
Reasoning & ReAct & .000 & .028 & .096 & .503 & .000 & .062 & .295 & .467 & .000 & .016 & .018 & .376 & .000 & .046 & .129 & .667 \\
\multirow{3}{*}{Iterative} & Iter-Retgen & .265 & .370 & .353 & .621 & .104 & .209 & .310 & .401 & .115 & .190 & .178 & .391 & .581 & .676 & .689 & .804 \\
& Self-RAG & .174 & .299 & .321 & .626 & .123 & .251 & .333 & .466 & .073 & .162 & .132 & .392 & .385 & .540 & .639 & .774 \\
& IRCoT & .006 & .087 & .399 & .631 & .001 & .085 & \textbf{.433} & \textbf{.488} & .001 & .056 & .210 & .416 & .016 & .169 & .674 & .800 \\
\hline
\multirow{6}{*}{Iterative} & FAIR-RAG 1 & \underline{.300} & \underline{.398} & .337 & \underline{.628} & \underline{.186} & \underline{.288} & .286 & .414 & \underline{.133} & \underline{.216} & .169 & \underline{.418} & \underline{.622} & \underline{.706} & .682 & \underline{.821} \\
& FAIR-RAG 2 & \underline{.335} & \underline{.447} & \underline{.397} & \underline{.689} & \underline{\textbf{.216}} & \underline{.325} & \underline{.341} & .458 & \underline{.168} & \underline{.253} & \underline{.214} & \underline{.465} & \underline{\textbf{.645}} & \underline{\textbf{.732}} & \underline{\textbf{.712}} & \underline{.837} \\
& FAIR-RAG 3 & \underline{.332} & \underline{.447} & \underline{\textbf{.404}} & \underline{\textbf{.697}} & \underline{.183} & \underline{.305} & \underline{.338} & .450 & \underline{\textbf{.178}} & \underline{\textbf{.267}} & \underline{.221} & \underline{.474} & \underline{.631} & \underline{.721} & \underline{.701} & \underline{.839} \\
& FAIR-RAG 4 & \underline{\textbf{.344}} & \underline{\textbf{.456}} & \underline{.401} & \underline{\textbf{.697}} & \underline{.209} & \underline{\textbf{.333}} & \underline{.357} & \underline{.486} & \underline{.175} & \underline{.266} & \underline{\textbf{.228}} & \underline{\textbf{.475}} & \underline{.644} & \underline{.728} & \underline{.710} & \underline{.837} \\
& FAIR-RAG 2 (Adpt.) & \underline{.331} & \underline{.436} & \underline{.384} & \underline{.673} & \underline{.183} & \underline{.296} & .313 & .444 & \underline{.158} & \underline{.241} & \underline{.207} & \underline{.447} & \underline{.640} & \underline{.722} & \underline{.704} & \underline{.828} \\
& FAIR-RAG 3 (Adpt.) & \underline{.338} & \underline{.453} & \underline{.399} & \underline{.694} & \underline{.206} & \underline{.320} & \underline{.350} & .452 & \underline{\textbf{.178}} & \underline{.264} & \underline{.222} & \underline{.472} & \underline{\textbf{.645}} & \underline{.731} & \underline{.710} & \underline{\textbf{.847}} \\
\hline
\end{tabular}
\caption{Main end-to-end performance comparison of FAIR-RAG against representative baselines across four diverse question-answering benchmarks. The evaluation covers three complex multi-hop datasets (HotpotQA, 2WikiMultiHopQA, Musique) and one open-domain factual dataset (TriviaQA). We report four metrics: Exact Match (EM), F1-Score, script-based Accuracy (ACC), and LLM-as-Judge Accuracy (ACC\textsubscript{LLM}). Our FAIR-RAG variants consistently achieve state-of-the-art performance, with the most significant improvements on the multi-hop tasks. The best-performing method for each metric is highlighted in \textbf{bold}. Underlined values denote results surpassing the Self-RAG and Iter-Retgen baselines. All scores are based on 1000 test/dev samples.}
\label{tab:main-results}
\end{table*}

Our best-performing model, \textbf{FAIR-RAG 3}, demonstrates leading performance within the paradigm of iterative and adaptive RAG. FAIR-RAG significantly outperforms strong representative baselines from its architectural class across all multi-hop benchmarks. \textbf{It is critical to note that all comparisons were conducted under a unified and controlled experimental setup, ensuring a fair and reproducible evaluation.} Within this rigorous framework, our results not only demonstrate a consistent advantage in our direct head-to-head comparisons but also \textbf{set a new state-of-the-art performance benchmark for this class of iterative methods on these datasets.} While we reference previously published results from comparable architectures~\cite{shao2023iter,asai2023self,jeong2024adaptiveraglearningadaptretrievalaugmented}, our primary claim of superiority is grounded in re-evaluating these methods under our standardized conditions to eliminate confounding variables. 

The most substantial gains are observed on the multi-hop reasoning benchmarks. On \textbf{HotpotQA}, our model achieves an F1 score of \textbf{0.453}, an absolute improvement of \textbf{+8.3 points} over the strongest baseline, Iter-Retgen (0.370). Similarly, on \textbf{2WikiMultiHopQA} and \textbf{Musique}, FAIR-RAG achieves F1 scores of \textbf{0.320} and \textbf{0.264}, respectively, outperforming the next-best method (Self-RAG) by a significant margin. These results strongly validate the efficacy of our core architectural contributions: the \textbf{Iterative Refinement Cycle} and the \textbf{Structured Evidence Assessment (SEA)}. These mechanisms empower FAIR-RAG to systematically deconstruct complex information needs, gather comprehensive evidence, and verify its sufficiency where single-pass or less structured iterative methods fall short.

Notably, FAIR-RAG also excels on the single-hop factual benchmark, \textbf{TriviaQA}, achieving an F1 score of \textbf{0.731}. This demonstrates that the framework's sophisticated reasoning machinery does not impose a penalty on simpler retrieval tasks and can effectively streamline its process, largely due to the initial \textbf{Adaptive Routing} module.

Comparing different variants of our model, we see a consistent performance increase from FAIR-RAG 1 to 4, indicating that the incremental enhancements contribute positively. The introduction of \textbf{Adaptive LLMs} further boosts performance across the board, confirming the benefits of dynamically allocating computational resources based on task complexity.

\begin{figure*}[t]
\centering
\includegraphics[width=\textwidth]{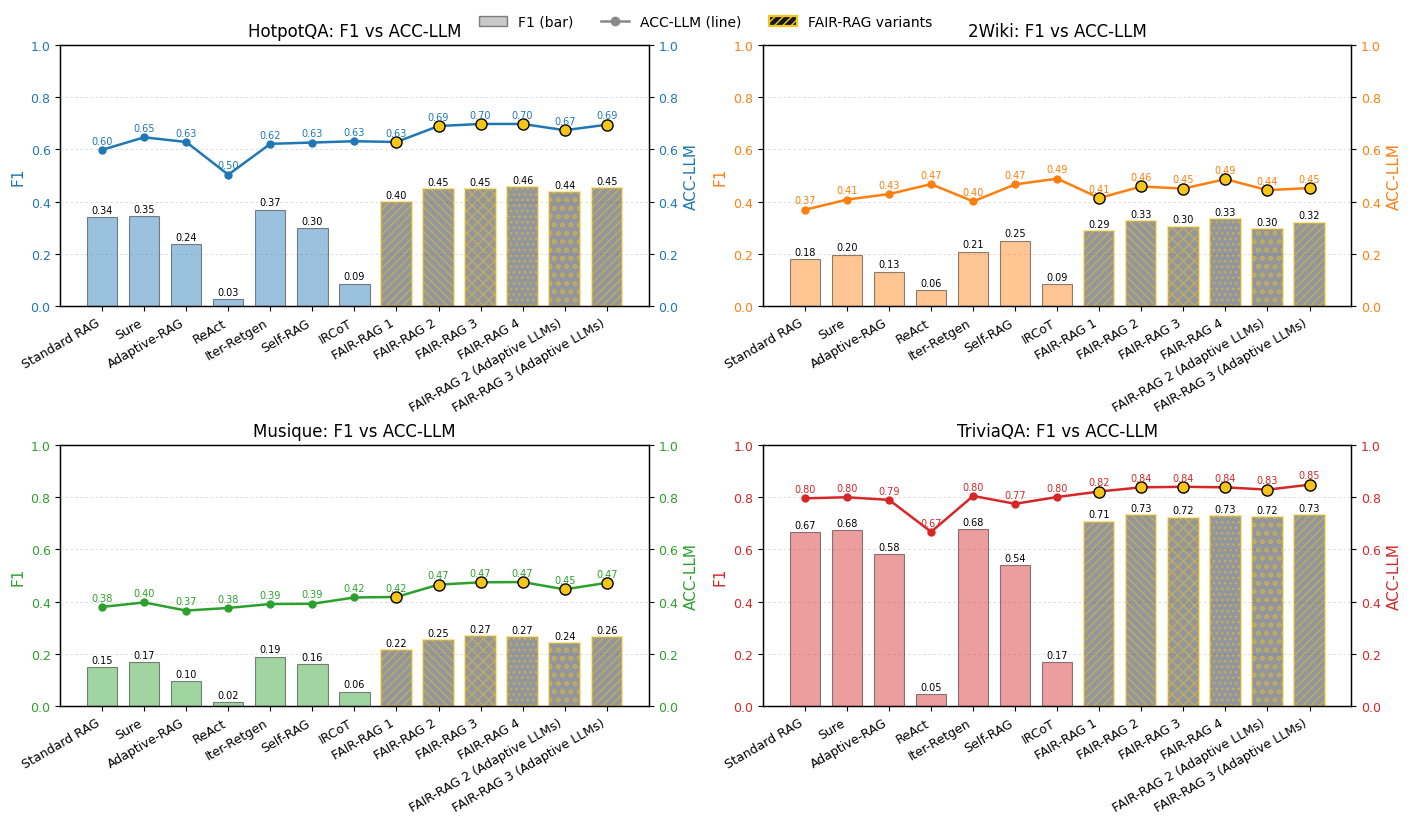}
\caption{Performance comparison of FAIR-RAG variants against baseline methods on four question-answering benchmarks. Each subplot displays the F1 score (bars, left axis) and the LLM-as-Judge Accuracy (ACC\textsubscript{LLM}, line, right axis). Our FAIR-RAG models are highlighted with a hatched pattern. The results consistently demonstrate the superiority of our framework, especially on complex multi-hop datasets (HotpotQA, 2WikiMultiHopQA, and MusiQue), where it significantly outperforms all baselines in F1 score while maintaining high semantic accuracy. All evaluations are conducted on 1000 samples from each benchmark's development set.}
\label{fig:performance-comparison}
\end{figure*}

To complement the tabular data, Figure~\ref{fig:performance-comparison} provides a visual comparison, plotting the F1 scores (bars) against the ACC\textsubscript{LLM} metric (lines) for all methods across the four benchmarks. The figure visually corroborates the superior performance of our FAIR-RAG framework, where its variants (indicated by the patterned bars) consistently achieve the highest F1 scores, particularly on the complex reasoning datasets of HotpotQA, 2WikiMultiHopQA, and Musique.

A key trend highlighted by the plots is the \textbf{strong positive correlation between the F1 score and the ACC\textsubscript{LLM}}. This suggests that the architectural improvements within FAIR-RAG, which enhance the accuracy of the Large Language Model's internal processing and decision-making, are directly responsible for the enhanced final answer accuracy. This relationship is particularly evident in the HotpotQA and 2WikiMultiHopQA results, where a noticeable uplift in the ACC\textsubscript{LLM} line for FAIR-RAG variants coincides with a significant increase in their corresponding F1 scores compared to the baselines. Thus, the visualizations not only confirm FAIR-RAG's state-of-the-art performance but also offer insight into the synergistic relationship between its internal reasoning accuracy and its final output quality.

\subsection{Further Analysis}

Beyond the main end-to-end results, we conduct a series of deeper analyses to deconstruct the sources of FAIR-RAG's performance. We perform a detailed component-wise evaluation to understand the contribution of each module, analyze the specific impact of the iterative refinement process, and present a qualitative case study to illustrate the framework in action. For these fine-grained assessments, we employ a sophisticated \textbf{LLM-as-Judge} methodology, using a powerful LLM to score the output of each internal module against a set of predefined criteria~\cite{Zheng2023Judging}. This task is guided by a structured prompt (see Appendix C for details).

\subsubsection{Component-wise Performance Analysis (Ablation Study)}
\label{sec:ablation}

To quantify the contribution of each key module within the FAIR-RAG pipeline, we conducted an ablation study using an LLM-as-Judge methodology on 1000 samples from each dataset. The results, summarized in Table~\ref{tab:ablation}, demonstrate the high functionality of each component, justifying its inclusion in the final architecture. For components like \textbf{Query Decomposition} and \textbf{Query Refinement}, the judge rated the output quality on a 1-to-5 Likert scale. The rating was based on a holistic assessment of criteria including \textbf{relevance} (how well it addresses the core need), \textbf{specificity} (how focused the query is), and \textbf{coverage} (whether it captures all necessary facets of the information gap). The average score across 1000 samples is reported. For \textbf{Evidence Filtering}, we report F1-Score based on the LLM's ability to correctly classify documents as relevant or irrelevant. For \textbf{SEA}, we report Accuracy based on its correctness in judging evidence sufficiency.

\begin{table*}[t]
\centering
\small
\setlength{\tabcolsep}{4pt}
\begin{tabular}{@{}lcccccccc@{}}
\hline
& \multicolumn{2}{c}{\textbf{Query Decomp.}} & \multicolumn{2}{c}{\textbf{Evidence Filter}} & \multicolumn{2}{c}{\textbf{SEA}} & \multicolumn{2}{c}{\textbf{Query Refine.}} \\
\cline{2-3} \cline{4-5} \cline{6-7} \cline{8-9}
\textbf{Dataset} & \textbf{Metric} & \textbf{Value} & \textbf{Metric} & \textbf{Value} & \textbf{Metric} & \textbf{Value} & \textbf{Metric} & \textbf{Value} \\
\hline
HotpotQA & Avg. Score & 4.19 & F1-Score & 67.3\% & Accuracy & 72.0\% & Avg. Score & 4.45 \\
2WikiMultiHopQA & Avg. Score & 4.12 & F1-Score & 55.5\% & Accuracy & 81.7\% & Avg. Score & 4.39 \\
Musique & Avg. Score & 4.10 & F1-Score & 68.5\% & Accuracy & 83.2\% & Avg. Score & 4.42 \\
TriviaQA & Avg. Score & 4.33 & F1-Score & 76.1\% & Accuracy & 54.4\% & Avg. Score & 4.52 \\
\hline
\end{tabular}
\caption{Component-level performance analysis of FAIR-RAG's core modules. This table isolates and evaluates the effectiveness of four key components: Query Decomposition, Evidence Filtering, Structured Evidence Assessment (SEA), and Query Refinement. The Avg. Score is based on a 1-to-5 Likert scale, while F1-Score and Accuracy are expressed as percentages. All scores averaged over 1000 samples.}
\label{tab:ablation}
\end{table*}

The analysis reveals several key insights:

\begin{itemize}
    \item \textbf{Query Decomposition \& Refinement are Highly Effective:} The initial \textbf{Query Decomposition} module achieves a high average quality score across all datasets, peaking at \textbf{4.33/5.0} on TriviaQA. The subsequent \textbf{Query Refinement} module scores even higher, with an average of \textbf{4.45/5.0} on HotpotQA. This validates that the LLM agents are proficient at both breaking down complex queries and intelligently generating new queries to fill information gaps identified by the SEA module.
    
    \item \textbf{Evidence Filtering Presents a Precision-Recall Trade-off:} The filtering module's performance varies, with F1 scores ranging from \textbf{55.47\%} on 2WikiMultiHopQA to \textbf{76.05\%} on TriviaQA. While the filter is effective at reducing context noise (precision), its aggressive nature can sometimes prune useful information (recall). This highlights a classic trade-off and presents a clear avenue for future optimization.
    
    \item \textbf{Structured Evidence Assessment (SEA) is a Challenging but Crucial Task:} The SEA module, which governs the iterative loop, demonstrates strong performance on complex multi-hop datasets, achieving an accuracy of \textbf{81.73\%} on 2WikiMultiHopQA and \textbf{83.19\%} on Musique. Its lower accuracy on TriviaQA (\textbf{54.38\%}) is expected, as the sufficiency decision for single-hop factual questions can be more ambiguous. These results confirm the module's value as the central control mechanism for the iterative process, proving highly reliable when it matters most.
\end{itemize}

It is important to note that a direct ablation study removing the SEA and Query Refinement modules and replacing them with a fixed-iteration loop was deliberately omitted. Such a stripped-down design, lacking adaptive control and explicit gap analysis, would cease to be FAIR-RAG and instead would conceptually approximate the architectural principles of existing iterative baselines. For instance, without a targeted query refinement strategy driven by SEA, the system would likely resort to generating new queries from the previous answer, a core mechanism in ITER-RETGEN~\cite{shao2023iter}. Furthermore, by removing its primary mechanism for evidence-aware self-correction, this simplified variant would represent a less sophisticated control strategy than that of models like Self-RAG~\cite{asai2023self}, which employ reflection at a more granular level.

Given that FAIR-RAG already demonstrates a significant performance margin over these strong baselines in our main experiments (see Table~\ref{tab:main-results}), our existing comparisons effectively serve as a validation of our adaptive, SEA-driven architecture versus these alternative control strategies. This confirms that the observed performance gains are attributable to the synergistic and intelligent control of the SEA and Query Refinement modules, rather than merely the brute-force effect of repeated iterations.

\subsubsection{Impact of Iterative Refinement}
\label{sec:iterative-impact}

A core hypothesis of this work is that iterative refinement improves answer quality for complex questions. We tested this by running the same set of questions with the maximum number of iterations ranging from 1 to 4. We then used an LLM-as-Judge to rank the resulting answers for each question. The results, along with efficiency metrics, are presented in Table~\ref{tab:iteration-impact}.

\begin{table*}[t]
\centering
\small
\setlength{\tabcolsep}{4pt}
\begin{tabular}{lccccc}
\hline
\textbf{Dataset} & \textbf{Max Iter.} & \textbf{Avg. Answer Rank} & \textbf{Improvement Rate} & \textbf{Avg. API Calls} & \textbf{Avg. Tokens/Query} \\
& & \textbf{(Lower is better)} & \textbf{(vs. Iter 1)} & \textbf{(\#)} & \\
\hline
\multirow{4}{*}{HotpotQA} 
& 1 & 2.73 & - & 4.97 & 9,787 \\
& 2 & \textbf{2.23} & \textbf{58.50\%} & 6.64 & 14,332 \\
& 3 & 2.38 & 57.00\% & 7.83 & 17,281 \\
& 4 & 2.66 & 57.00\% & 9.01 & 20,299 \\
\hline
\multirow{4}{*}{2WikiMultiHopQA} 
& 1 & 3.08 & - & 4.99 & 9,823 \\
& 2 & 2.31 & 69.30\% & 7.10 & 15,413 \\
& 3 & \textbf{2.18} & \textbf{70.90\%} & 8.79 & 19,812 \\
& 4 & 2.43 & 67.30\% & 10.14 & 23,231 \\
\hline
\multirow{4}{*}{Musique} 
& 1 & 2.89 & - & 4.98 & 9,613 \\
& 2 & \textbf{2.20} & 63.40\% & 7.25 & 15,688 \\
& 3 & 2.28 & \textbf{63.70\%} & 9.01 & 20,162 \\
& 4 & 2.63 & 61.20\% & 10.56 & 24,218 \\
\hline
\multirow{4}{*}{TriviaQA} 
& 1 & \textbf{1.83} & - & 4.97 & 9,572 \\
& 2 & 2.08 & 28.50\% & 5.97 & 12,071 \\
& 3 & 2.70 & 27.20\% & 6.76 & 14,003 \\
& 4 & 3.39 & 27.20\% & 7.25 & 15,186 \\
\hline
\end{tabular}
\caption{Ablation study on the impact of the maximum number of refinement iterations on answer quality versus computational cost. The results reveal a clear point of diminishing returns. For complex multi-hop datasets, optimal performance (lowest Avg. Answer Rank) is achieved at 2 or 3 iterations, after which quality degrades while costs (API Calls, Tokens) continue to increase linearly. Conversely, for the simpler fact-based TriviaQA, any iteration beyond the first proves detrimental. This analysis empirically justifies our framework's default setting of a three-iteration maximum.}
\label{tab:iteration-impact}
\end{table*}

The data reveals a clear and consistent pattern across the three multi-hop datasets (\textbf{HotpotQA, 2WikiMultiHopQA, and Musique}):

\begin{itemize}
    \item \textbf{Optimal Performance at 2-3 Iterations:} Moving from one to two iterations yields a substantial improvement in answer quality. On 2WikiMultiHopQA, the average quality rank improves from \textbf{3.08} to \textbf{2.31}. The peak performance is generally observed at either the second or third iteration (e.g., an average rank of \textbf{2.18} at iteration 3 for 2WikiMultiHopQA). This is accompanied by a high \textbf{Improvement Rate}, with the 2- or 3-iteration answer being judged superior to the 1-iteration answer in approximately \textbf{70\%} of cases for 2WikiMultiHopQA.
    
    \item \textbf{Diminishing Returns:} A fourth iteration consistently leads to a degradation in average answer quality across all complex datasets. This suggests a point of diminishing returns where an additional retrieval cycle is more likely to introduce noisy or tangentially related information that complicates the final synthesis step.
    
    \item \textbf{Cost-Benefit Analysis:} Each iteration adds a considerable number of API calls and tokens, increasing both latency and computational cost. The optimal balance of 2-3 iterations provides the best balance between answer quality and resource consumption.
\end{itemize}

Conversely, on the simpler \textbf{TriviaQA} dataset, the quality rank degrades with each additional iteration, confirming that for single-hop queries, the initial retrieval is generally sufficient, and further iterations are unnecessary and even detrimental. This analysis confirms that iteration is crucial for complex reasoning, but an unrestrained number of iterations is suboptimal. Our framework's default setting of a maximum of 3 iterations is thereby empirically justified as an effective balance between performance and efficiency.

\subsubsection{A Complex Case Study: Comparative Multi-Hop Reasoning}

To demonstrate the unique advantages of the FAIR-RAG architecture over other advanced RAG frameworks, we analyze a hybrid \textbf{comparative, multi-hop query}. This type of query is particularly challenging because it requires the system to conduct two parallel lines of multi-hop reasoning simultaneously and then synthesize the results.

\begin{figure*}[t]
\centering
\includegraphics[width=\textwidth]{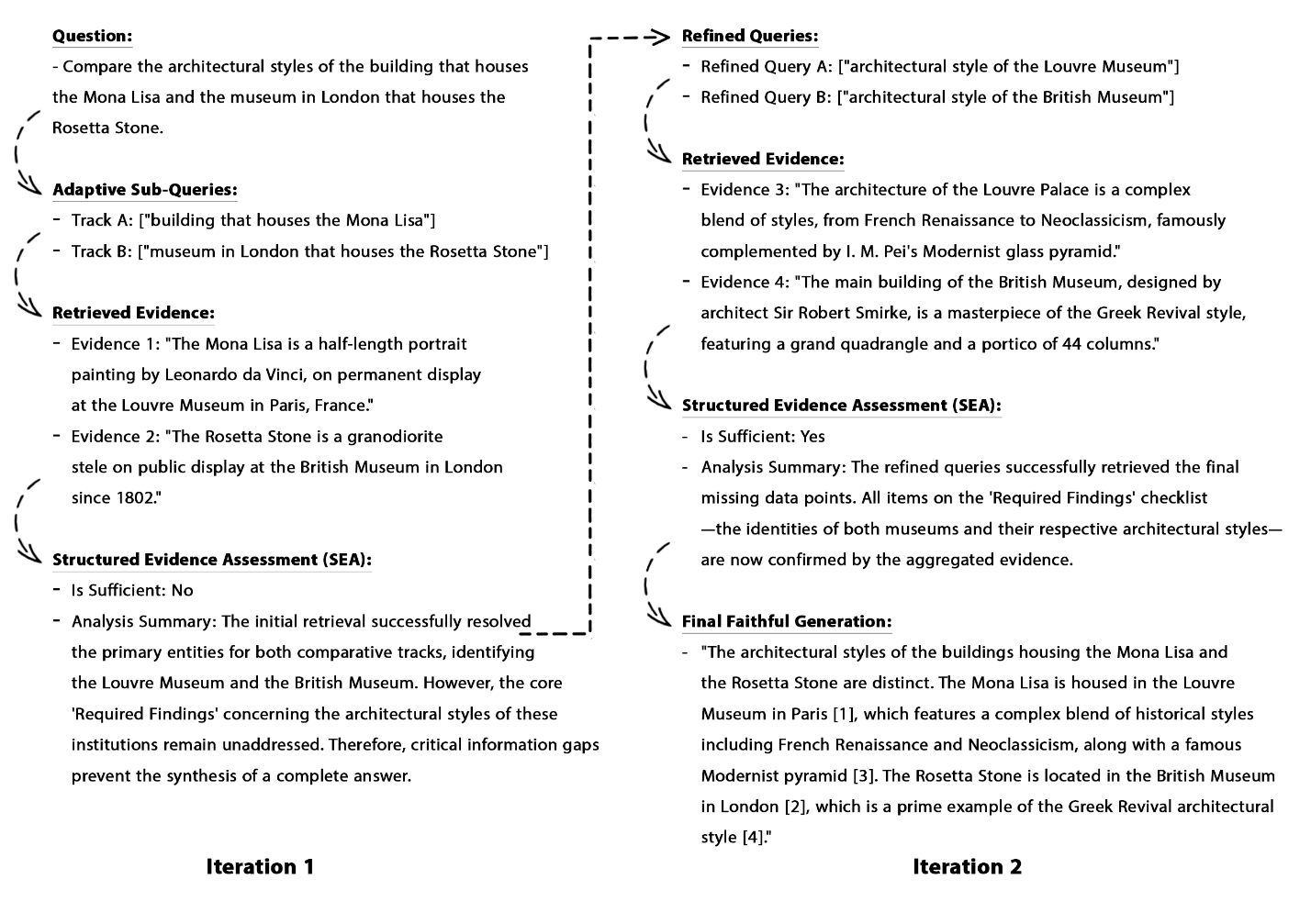}
\caption{A qualitative case study demonstrating FAIR-RAG's two-iteration process for a complex comparative query. In Iteration 1, the system decomposes the query and retrieves initial evidence identifying the museums but lacking the required architectural styles. The Structured Evidence Assessment (SEA) module correctly identifies this gap (Is Sufficient: No), triggering a second iteration. In Iteration 2, the system generates Refined Queries targeting the missing information, successfully retrieves the necessary evidence, and confirms sufficiency. The process culminates in a Final Faithful Generation that synthesizes evidence from both iterations into a comprehensive answer.}
\label{fig:case-study}
\end{figure*}

\textbf{The query is:} \textbf{``Compare the architectural styles of the building that houses the \emph{Mona Lisa} and the museum in London that houses the \emph{Rosetta Stone}.''}

\paragraph{Standard RAG Failure:}
A standard RAG system would treat this complex comparative query as a single, semantically overloaded search vector. This unfocused approach is highly likely to fail for two primary reasons: First, it would struggle to simultaneously retrieve relevant, detailed documents for both distinct lines of inquiry (the Louvre and the British Museum). It might retrieve a general document about the Mona Lisa that lacks architectural details, or miss one of the entities entirely. Second, and more fundamentally, standard RAG lacks the procedural logic to deconstruct the query, pursue two parallel reasoning paths, and then synthesize the findings into a coherent comparison. It relies on finding a single document that already compares the two museums' architecture, which is highly improbable. Consequently, a standard RAG would likely produce a disjointed answer focusing on only one of the entities, or fail entirely.

\paragraph{Why this query is difficult for other advanced RAGs:}
\begin{itemize}
    \item An \textbf{ITER-RETGEN} system, due to its inherently single-threaded nature, might successfully follow one reasoning path (e.g., find the Louvre, then its style). However, it lacks the mechanism to manage a parallel track simultaneously, causing it to lose the context of the second entity (the British Museum) and fail to produce a coherent comparison.
    \item An \textbf{Adaptive-RAG} framework may correctly identify the query as ``complex.'' However, it typically lacks a structured, multi-track decomposition process. Without the ability to systematically pursue both information threads in parallel before synthesis, its adaptive strategy remains insufficient for true comparative reasoning.
\end{itemize}

\paragraph{FAIR-RAG in Action:}
\begin{itemize}
    \item \textbf{Iteration 1: Semantic Decomposition \& Parallel Initial Retrieval}
    \begin{itemize}
        \item \textbf{Adaptive Sub-Queries:} FAIR-RAG's first action is to decompose the comparative query into two distinct, parallel investigative tracks:
        \begin{itemize}
            \item Track A: [``building that houses the Mona Lisa'']
            \item Track B: [``museum in London that houses the Rosetta Stone'']
        \end{itemize}
        \item \textbf{Retrieved Evidence:} The system retrieves evidence for both tracks concurrently:
        \begin{itemize}
            \item Evidence A: ``The Mona Lisa is a half-length portrait painting by Leonardo da Vinci, on permanent display at the \textbf{Louvre Museum} in Paris, France.''
            \item Evidence B: ``The Rosetta Stone is a granodiorite stele on public display at the \textbf{British Museum} in London since 1802.''
        \end{itemize}
        \item \textbf{Structured Evidence Assessment (SEA):}
        \begin{itemize}
            \item \textbf{is\_sufficient:} \textbf{`No'}
            \item \textbf{analysis\_summary:} The initial retrieval successfully resolved the primary entities for both comparative tracks, identifying the Louvre Museum and the British Museum. However, the core `Required Findings' concerning the architectural styles of these institutions remain unaddressed. Therefore, critical information gaps prevent the synthesis of a complete answer.
        \end{itemize}
    \end{itemize}
    
    \item \textbf{Iteration 2: Parallel Query Refinement \& Evidence Completion}
    \begin{itemize}
        \item \textbf{Refined Queries:} The refinement module now uses the entities identified in Iteration 1 to generate new, highly-focused queries for each track:
        \begin{itemize}
            \item Refined Query A: [``architectural style of the Louvre Museum'']
            \item Refined Query B: [``architectural style of the British Museum'']
        \end{itemize}
        \item \textbf{Retrieved Evidence:} The new targeted queries retrieve the final missing pieces of information:
        \begin{itemize}
            \item Evidence A: ``The architecture of the Louvre Palace is a complex blend of styles, from French Renaissance to Neoclassicism, famously complemented by I. M. Pei's Modernist glass pyramid.''
            \item Evidence B: ``The main building of the British Museum, designed by architect Sir Robert Smirke, is a masterpiece of the \textbf{Greek Revival} style, featuring a grand quadrangle and a portico of 44 columns.''
        \end{itemize}
        \item \textbf{Structured Evidence Assessment (SEA):}
        \begin{itemize}
            \item \textbf{is\_sufficient: `Yes'}
            \item \textbf{analysis\_summary:} The refined queries successfully retrieved the final missing data points. All items on the `Required Findings' checklist---the identities of both museums and their respective architectural styles---are now confirmed by the aggregated evidence. No information gaps remain, enabling a direct, evidence-grounded comparison.
        \end{itemize}
    \end{itemize}
    
    \item \textbf{Final Faithful Generation:}
    The system synthesizes the evidence from \textbf{both parallel tracks} into a single, structured, and fully-grounded comparative answer:
    
    ``The architectural styles of the buildings housing the Mona Lisa and the Rosetta Stone are distinct. The Mona Lisa is housed in the \textbf{Louvre Museum} in Paris [1], which features a complex blend of historical styles including \textbf{French Renaissance and Neoclassicism}, along with a famous \textbf{Modernist} pyramid [3]. The Rosetta Stone is located in the \textbf{British Museum} in London [2], which is a prime example of the \textbf{Greek Revival} architectural style [4].''
\end{itemize}

This case study highlights FAIR-RAG's key architectural advantage: its ability to systematically \textbf{decompose} a complex goal into parallel sub-problems and then apply \textbf{iterative refinement} to solve each one before synthesizing a comprehensive final answer. This structured, multi-threaded reasoning process is what enables it to succeed where other iterative or adaptive frameworks may fail.

\subsubsection{Failure Mode Analysis}
\label{sec:failure-analysis}

To foster transparency and identify avenues for future improvement, we conducted a systematic failure mode analysis on a sample of 200 unique error instances drawn equally from our four benchmark datasets: TriviaQA, MuSiQue, HotpotQA, and 2WikiMultihopQA. This analysis employed a hybrid methodology, combining LLM-based categorization with human expert validation to ensure accuracy and depth. Our taxonomy distinguishes between two fundamental sources of error: (1) \textbf{Component-Level Failures}, which stem from the inherent limitations of the underlying modules (i.e., the retriever and the generator LLM), and (2) \textbf{Architectural Failures}, which are specific to the decision-making logic of the FAIR-RAG framework itself (i.e., Query Decomposition, Filtering, Refinement, and SEA).

\begin{figure}[t]
\centering
\includegraphics[width=0.95\columnwidth]{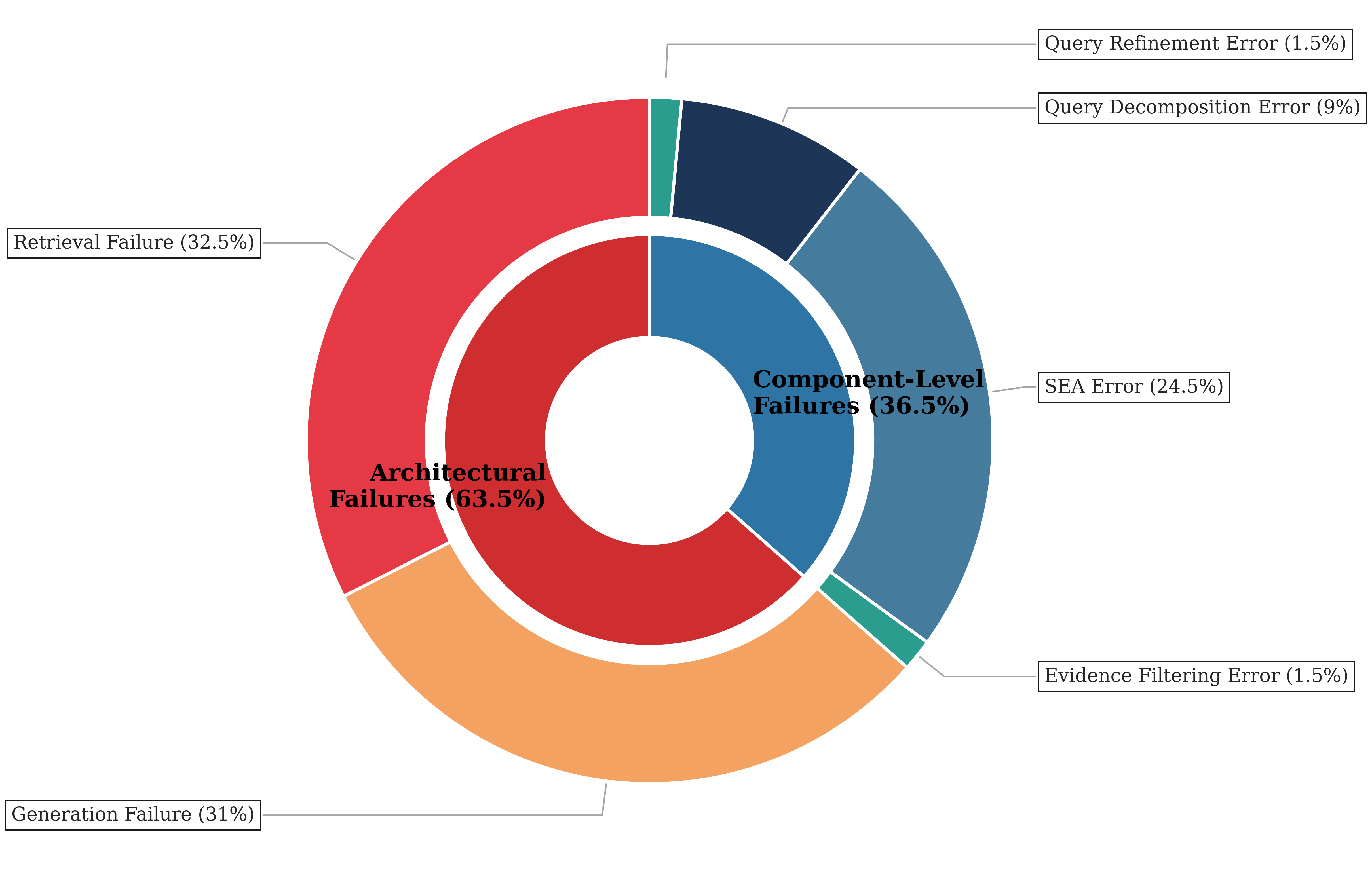}
\caption{Aggregate Distribution of Failure Sources. Analysis of 200 error samples reveals a primary split between Component-Level Failures (63.5\%) and Architectural Failures (36.5\%). While architectural logic offers direct avenues for refinement, the majority of errors stem from the inherent limitations of the foundational retrieval and generation models, identifying them as the principal bottleneck for the FAIR-RAG system.}
\label{fig:failure-sources}
\end{figure}

This distinction is critical for understanding the system's bottlenecks. As shown in Figure~\ref{fig:failure-sources}, a significant majority of errors (63.5\%) are Component-Level, originating from the foundational tools our system is built upon. The remaining 36.5\% are Architectural, offering direct targets for refining FAIR-RAG's internal logic. This distribution underscores a key insight: while FAIR-RAG's iterative process is designed to mitigate the weaknesses of its components, the performance of these base components remains the primary limiting factor in overall system accuracy.

\paragraph{1. Component-Level Failures (63.5\% of Errors): The Foundational Bottleneck}

These errors are not caused by FAIR-RAG's reasoning process but by the fundamental limitations of the tools it orchestrates.

\begin{itemize}
    \item \textbf{Retrieval Failure (32.5\%):} This was the single largest source of error across all datasets. The retriever's inability to surface critical documents is a major obstacle, particularly for queries requiring highly specific, long-tail factual knowledge. The primary root cause was identified as Knowledge Base Gaps, where the required information was simply absent from the corpus. This was especially pronounced in TriviaQA, where nearly half of all failures were retrieval-related due to the dataset's reliance on obscure facts.
    
    \item \textbf{Generation Failure (31.0\%):} In these cases, the correct evidence was successfully identified and passed to the final generator, which still produced a flawed answer. This category represents a significant challenge across all datasets, highlighting the inherent faithfulness problem in modern LLMs. Common failure subtypes included: (i) Incorrect Entity Selection, where the model chose the wrong entity from a list of candidates in the evidence; (ii) Flawed Logical Inference, especially in comparative questions (e.g., ``who is younger?''); and (iii) Misinterpretation of Question Granularity, such as providing a specific year (``1922'') when a decade (``1920s'') was requested.
\end{itemize}

\paragraph{2. Architectural Failures (36.5\% of Errors): Refining the FAIR-RAG Logic}

These errors are directly attributable to FAIR-RAG's internal decision-making modules and represent the most direct opportunities for improving our framework.

\begin{itemize}
    \item \textbf{SEA Error (24.5\%):} As the ``brain'' of the iterative process, failures in the Strategic Evidence Assessment module are particularly impactful. These errors were significantly more prevalent in complex, multi-hop datasets like MuSiQue, HotpotQA, and 2WikiMultihopQA. The most common subtypes were: (i) Faulty Analysis of Evidence, where the SEA module failed to make a correct logical inference from the provided documents (e.g., misinterpreting complex genealogical relationships); and (ii) Premature Sufficiency Judgment, where the logic incorrectly concluded that the gathered evidence was complete, thus halting the refinement loop too early.
    
    \item \textbf{Query Logic Failures (12.0\% combined):} This group includes errors from the initial query processing stages. Query Decomposition Errors (9.0\%) were the most common, often stemming from an inability to challenge flawed premises within the user's question (e.g., processing a query with a historical anachronism) or generating overly broad sub-queries. Query Refinement (1.5\%) and Evidence Filtering Errors (1.5\%) were exceedingly rare, suggesting that these architectural components are relatively robust.
\end{itemize}

\paragraph{Dataset-Specific Error Distributions}

\begin{figure}[t]
\centering
\includegraphics[width=0.95\columnwidth]{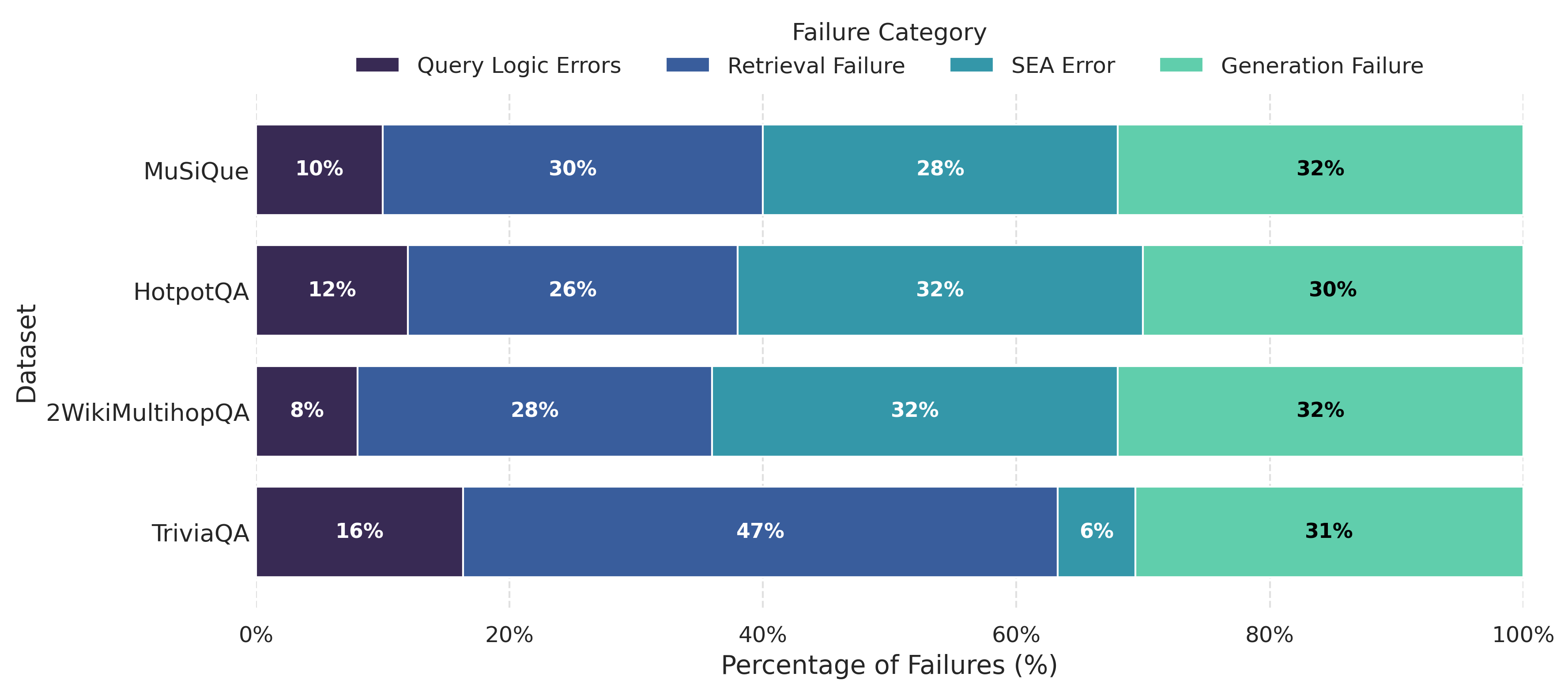}
\caption{Task-Dependent Distribution of Failure Modes. The analysis highlights a strong correlation between task complexity and the primary failure bottleneck. For the factoid-centric TriviaQA, Retrieval Failures are dominant (47\%). Conversely, for complex multi-hop reasoning datasets like MuSiQue, HotpotQA, and 2WikiMultihopQA, the burden shifts towards reasoning, with SEA Errors becoming a major failure category (28-32\%). This trend validates the critical role of the strategic reasoning component (SEA) for successfully navigating multi-step queries.}
\label{fig:failure-modes-dist}
\end{figure}

While the aggregate view is informative, Figure~\ref{fig:failure-modes-dist} reveals that the distribution of failure modes is highly dependent on the nature of the task. For fact-based, single-hop datasets like TriviaQA, failures are overwhelmingly concentrated in the Retrieval stage (47\%). However, as query complexity increases in multi-hop datasets (MuSiQue, HotpotQA, 2WikiMultihopQA), the burden shifts. In these cases, SEA Errors become significantly more prominent, accounting for 28-32\% of all failures. This demonstrates that for complex reasoning tasks, the primary challenge moves beyond simply finding information to correctly reasoning about and managing it. This trend strongly validates the necessity of a sophisticated strategic control module like SEA in advanced RAG systems.

\section{Conclusion}

In this paper, we introduced \textbf{FAIR-RAG}, a novel, agentic framework designed to address a key limitation of existing Retrieval-Augmented Generation systems: their unreliability in handling complex, multi-hop queries. By architecting an evidence-driven, iterative process, FAIR-RAG advances beyond the static ``retrieve-then-read'' paradigm. Our core contributions---\textbf{Adaptive Routing}, the \textbf{Iterative Refinement Cycle}, and the analytical gating mechanism of the \textbf{Structured Evidence Assessment (SEA)} module---work in synergy to progressively build and validate a comprehensive context before generation. The SEA's ability to systematically deconstruct queries and identify information gaps provides a precise, actionable signal that directly guides the query refinement process.

Our extensive experiments demonstrate that FAIR-RAG's structured approach achieves leading performance among comparable iterative and adaptive RAG architectures across challenging multi-hop QA datasets like HotpotQA and 2WikiMultiHopQA. These results empirically validate our central hypothesis: that a procedural, multi-stage workflow with explicit evidence assessment is essential for achieving high accuracy and faithfulness in knowledge-intensive tasks.

\subsection{Limitations}

Despite its strong performance, FAIR-RAG presents several inherent trade-offs and limitations that warrant discussion:

\begin{itemize}
    \item \textbf{Dependency on LLM Reasoning Fidelity and Prompt Engineering:} The performance of FAIR-RAG's modular agents (e.g., query refinement, SEA) is inherently bound by two key factors: the reasoning capabilities of the underlying LLMs and the meticulous design of the prompts that guide them. While our structured prompting methodology---which incorporates clear instructions, illustrative examples, and scaffolding techniques---is designed to ensure consistency and robustness, the system's effectiveness remains sensitive to both the choice of the backbone model and the specific phrasing of the prompts. This dual dependency is an intrinsic limitation of current LLM-based agentic systems, where performance gains are often tightly coupled with advancements in both model architecture and sophisticated prompt engineering.
    
    \item \textbf{Comprehensiveness vs. Efficiency Trade-off:} The iterative nature of FAIR-RAG, which is key to its high accuracy on complex queries, introduces a natural trade-off with efficiency. As shown in our analysis (Table~\ref{tab:iteration-impact}), each refinement cycle increases overall latency and computational cost, making it more expensive than single-shot RAG methods.
    
    \item \textbf{Potential for Error Propagation:} As a multi-stage pipeline, errors in early stages can cascade. The SEA module, in particular, represents a critical point of failure. An erroneous sufficiency judgment---either a false positive that terminates the loop prematurely or a false negative that extends it unnecessarily---can lead the evidence-gathering process astray.
    
    \item \textbf{Fixed Iteration Policy:} The maximum of three iterations is an empirically derived heuristic that balances performance and cost. However, a fixed limit lacks the flexibility to adapt to the varying complexity of individual queries. Our results show this is a robust average but may not be optimal for every specific case.
\end{itemize}

\subsection{Future Work}

The limitations of our current work open up several promising directions for future research to create more efficient and adaptive systems:

\begin{itemize}
    \item \textbf{Distilling Task-Specific Expert Models:} To mitigate the reliance on expensive, general-purpose LLMs, a promising direction is to fine-tune or distill smaller, specialized models for each core task~\cite{hinton2015distillingknowledgeneuralnetwork}. Creating dedicated expert models for query refinement or evidence assessment could lead to a system that is faster, cheaper, and more robust.
    
    \item \textbf{Learning a Dynamic Control Policy:} The iterative process can be framed as a sequential decision-making problem. We propose exploring Reinforcement Learning (RL) to train a policy network that learns to dynamically control the workflow~\cite{schick2023toolformerlanguagemodelsteach}. At each step, this agent could decide whether to retrieve more information, refine the query, or proceed to generation, replacing the fixed-loop structure with a far more efficient and adaptive strategy.
    
    \item \textbf{Extension to Multimodal Reasoning:} The current FAIR-RAG framework operates exclusively on textual data. A natural and impactful extension would be to adapt its core principles of decomposition, iterative refinement, and structured assessment to handle queries over multimodal knowledge bases that include tables, images, and structured data, creating a more versatile and comprehensive question-answering system~\cite{alayrac2022flamingovisuallanguagemodel}.
\end{itemize}

\bibliography{anthology,custom,references}
\bibliographystyle{acl_natbib}

\appendix

\section{Implementation Details and Hyperparameters}
\label{sec:appendix-implementation}

To ensure full reproducibility of our experimental results, this section provides a comprehensive overview of the models, configurations, and hyperparameters used in our study. All experiments were conducted within the \textbf{FlashRAG} framework, a standardized library for RAG research. This ensures that the underlying implementation details, such as data loading and prompt templating, remain consistent across all compared methods. The source code and default configurations can be found on the official project repository: \url{https://github.com/ruc-nlpir/flashrag}.

The specific configurations for our experiments are detailed in Table~\ref{tab:hyperparameters} below.

\begin{table}[h]
\centering
\small
\begin{tabular}{ll}
\hline
\textbf{Category} & \textbf{Parameter \& Value} \\
\hline
\multicolumn{2}{l}{\textbf{General Framework}} \\
& Base Framework: FlashRAG \\
& Max Iterations: 3 \\
\hline
\multicolumn{2}{l}{\textbf{Retriever Configuration}} \\
& Model: e5-base-v2 \\
& Documents Retrieved (top\_k): 5 \\
& Index Type: Faiss (IndexFlatIP) \\
& Pooling Method: mean \\
& Query Max Length: 128 \\
\hline
\multicolumn{2}{l}{\textbf{Generator Configuration}} \\
& Baseline: Llama-3-8B-Instruct \\
& Self-RAG: selfrag-llama-7b \\
& FAIR-RAG: Llama-3-8B-Instruct \\
& Max New Tokens: 1024 \\
& Max Input Length: 8000 \\
\hline
\multicolumn{2}{l}{\textbf{Evaluation Settings}} \\
& Metrics: em, f1, acc \\
& Random Sample: True \\
\hline
\end{tabular}
\caption{Hyperparameter and Model Configuration Details. Unless specified otherwise, all hyperparameters such as temperature, top\_p, and sequence lengths were kept at the default values provided by the FlashRAG framework to ensure a fair and controlled comparison across all tested methodologies.}
\label{tab:hyperparameters}
\end{table}

\section{Full Prompts for the FAIR-RAG Pipeline}
\label{sec:appendix-prompts}

This appendix reproduces the exact prompts that guide the behavior of the specialized agents within the FAIR-RAG pipeline.

\subsection{Query Validation and Dynamic Model Selection}

The following prompt is used by the initial agent to validate the user's query for clarity and safety, and to select the most appropriate execution model (e.g., simple RAG vs. full agentic pipeline).

\begin{lstlisting}[basicstyle=\scriptsize\ttfamily]
PROMPT = """

**Situation:** A user has submitted a question to a Question Answering System that uses different processing strategies based on query complexity.

**Intent:**

Analyze the user's question to determine the optimal processing strategy required to generate the most accurate answer. The strategies are: Factual Retrieval (SMALL), Information Synthesis (LARGE), or Multi-Step Deduction (REASONER).

**Scaffolding:**

You are a highly-calibrated query analysis agent. Your task is to classify the user's question into one of the three categories below based on the cognitive process required to answer it. After "Selected Label:", output ONLY the exact label.

- **"SMALL" (Factual Retrieval):**

    - **Process:** Requires finding a single, self-contained fact. The answer is typically explicit and doesn't need much context.

    - **Example:** "When was the Eiffel Tower completed?" or "Who is the CEO of NVIDIA?"

- **"LARGE" (Information Synthesis):**

    - **Process:** Requires combining, summarizing, or comparing information from one or more documents to form a coherent, explanatory answer. This is the appropriate choice for most standard, open-ended questions.

    - **Example:** "What is the difference between nuclear fission and fusion?" or "Explain the impact of the printing press on the Renaissance."

- **"REASONER" (Multi-Step Deduction):**

    - **Process:** The answer is not explicitly stated and must be inferred by chaining multiple pieces of information together (multi-hop reasoning) or by performing calculations.

    - **Example:** "What is the hometown of the director of the movie that starred Tom Hanks and was released in 1994?" or "If a car travels 150 km in 2 hours, what is its average speed in meters per second?"

**Analytical Guidelines for Classification:**

1. "SMALL" (Factual Extraction):

     - Cognitive Task: Find and extract a specific, named entity or a short, self-contained fact (e.g., a name, date, location, number, or title).

     - Key Indicators: The question can be answered with a single piece of information, often a proper noun or a specific value. It typically starts with "Who," "When," "Where," or "What is the name of..."

     - Example Questions: "Who directed the movie Inception?" or "What year did the Berlin Wall fall?"

     - Decision Rule: Classify as SMALL if the expected answer is a concise, singular fact that requires no further explanation or combination of information.

2. "LARGE" (Information Synthesis & Elaboration):

     - Cognitive Task: Gather, combine, and summarize information from one or multiple sources to form a cohesive, descriptive answer. This involves explaining concepts, comparing entities, or describing processes.

     - Key Indicators: The question asks for an explanation, description, comparison, or summary. It often contains keywords like "Explain," "Describe," "Compare," "What is the difference between," or "Summarize the impact of..."

     - Example Questions: "What are the main differences between crocodiles and alligators?" or "Explain the primary causes of the Industrial Revolution."

     - Decision Rule: Classify as LARGE if the expected answer is a paragraph or a detailed sentence that combines multiple facts into a comprehensive explanation.

3. "REASONER" (Logical Inference & Multi-Step Deduction):

     - Cognitive Task: Connect multiple, separate pieces of information to infer a new fact that is not explicitly stated in any single document. This often involves a chain of logic or a sequence of dependent questions.

     - Key Indicators: The question requires finding an intermediate answer to proceed to the next step. It often involves relationships between different entities or requires a calculation.

     - Example Questions: "What was the nationality of the lead actress in the movie directed by the person who made Titanic?" or "Which team won the FIFA World Cup in the year the lead singer of Queen was born?"

     - Decision Rule: Classify as REASONER if the answer cannot be found directly but must be constructed by first finding Fact A, then using Fact A to find Fact B.

**User Question:** "{user_query}"

**Constraints:**

- Respond with ONLY one of the three labels: SMALL, LARGE, or REASONER.

- Do not provide any explanations or additional text.

- The label MUST be on a new line after "Selected Label:".

**Output:**

Selected Label:

"""
\end{lstlisting}

\subsection{Query Decomposition}

This prompt instructs the decomposition agent to break down a complex, multi-faceted user query into a set of simpler, semantically distinct sub-queries for parallel or sequential retrieval.

\begin{lstlisting}[basicstyle=\scriptsize\ttfamily]
PROMPT = """

**Situation:** You are an expert query analyst for a general knowledge Question-Answering system. A user has asked a question that might be complex, comparative, or multi-faceted. Your task is to decompose this question into a set of precise, meaningful, and distinct sub-queries to ensure the retrieval system can find comprehensive and accurate evidence from a database.

**Intent:** Decompose the original user question into its core semantic components. Transform these components into short, keyword-rich, and meaningful search phrases in English. The goal is to generate queries that, when searched, will collectively cover all aspects of the original question.

**Scaffolding:**

First, understand the principles of effective decomposition:

1. **Identify Distinct Concepts:** Separate the main subjects, actions, conditions, and comparisons in the query.

2. **Use Synonyms & Related Terms:** Think about different ways a concept might be phrased in the database (e.g., "interaction" can be searched as "relationship" or "cooperation").

3. **Create Meaningful Phrases:** Instead of single keywords, generate short phrases that preserve the context of the sub-question.

4. **Cover All Angles:** Ensure every part of the original question is represented by at least one sub-query.

Now, study the following example carefully to understand how to apply these principles.

--- EXAMPLE ---

**Original User Query:** "What was Albert Einstein's view on quantum mechanics and how did he interact with Niels Bohr about it?"

**Rationale/Analysis (This is your thought process):**

The query has two main, distinct parts:

1. Einstein's **opinion/view** about quantum mechanics.

2. Einstein's **interaction** with Niels Bohr on the topic.

A good search needs to find evidence for both aspects separately. Simply searching for "Einstein quantum mechanics" might not retrieve documents that specifically discuss his "view" or "interaction". Therefore, I should create targeted queries for each concept.

**Optimized Queries (Output):**

- Einstein's opinion on quantum mechanics
- Einstein Bohr debates on quantum theory
- Collaboration between Einstein and Bohr
- Einstein's criticism of quantum mechanics

--- END OF EXAMPLE ---

Now, apply this exact methodology to decompose the following query.

**User Query:** "{user_query}"

**Constraints:**

- The output must be a list of meaningful search phrases.

- Each phrase must be on a new line, prefixed with a hyphen (-).

- Queries must be in English.

- Generate an optimized list of 1 to 4 sub-queries. Create ONLY as many as are **truly necessary** to cover all aspects of the original question.

**Output:** (just write Optimized Queries and do not explain any more and do not say "Here are the optimized queries:" or something like that.)

Optimized Queries: (A list of optimized queries)

"""
\end{lstlisting}

\subsection{Evidence Filtering}

This prompt guides the filtering agent to assess the relevance and quality of the retrieved evidence chunks against a given (sub-)query, discarding irrelevant, redundant, or low-quality information.

\begin{lstlisting}[basicstyle=\scriptsize\ttfamily]
PROMPT = """

You are filtering retrieved documents for a question-answering system. Your goal is to KEEP all documents that could contribute to answering the query.

**IMPORTANT PRINCIPLES:**

1. BE INCLUSIVE: When in doubt, KEEP the document.

2. A document is useful if it contains factual information about the entities/topics in the query.

3. Even partial information is valuable (e.g., a document about Terry Richardson without birthdate is still useful for a query about his age).

4. Documents are only "Not Useful" if they are about COMPLETELY DIFFERENT entities or topics.

**Original User Query:** "{original_query}"

**Retrieved Documents (Batch {batch_number}):**

{numbered_candidates_text_for_prompt}

**TASK:** Identify ONLY documents that are completely irrelevant. A document is irrelevant ONLY if:

- It's about a different person/entity with a similar name (e.g., Tony Richardson vs Terry Richardson).

- It's about a completely unrelated topic.

- It contains no information about any entity mentioned in the query.

**OUTPUT FORMAT:**

- List ONLY the document IDs that should be removed

- If all documents are potentially useful, output: None

- Format: [doc_X], [doc_Y] or None

**You MUST KEEP documents that have:**

- Documents about the correct person/entity, even without specific dates/facts.

- Biographical information (birth dates, career details, achievements).

- Relationships or connections between queried entities.

- Specific facts relevant to the query type (dates for temporal queries, attributes for comparisons).

- Contextual information that helps understand the entities.

**Examples of what to REMOVE:**

- Documents about different people with similar names.

- Documents about unrelated topics.

- Duplicate documents (keep the most informative version).

**Output:** (A list of **irrelevant** temporary document IDs, or "None")

Unhelpful Document IDs:

"""
\end{lstlisting}

\subsection{Structured Evidence Assessment (SEA)}

The Structured Evidence Assessment (SEA) agent uses the following prompt to analyse the evidence and determine if the gathered and filtered evidence is adequate to form a complete and faithful answer to the user's query.

\begin{lstlisting}[basicstyle=\scriptsize\ttfamily]
PROMPT = """

**Role:** You are a Strategic Intelligence Analyst. Your mission is to determine if the provided evidence is sufficient to accurately answer the user's question by following a sequential analysis.

**Core Mission:** Your entire process must be question-centric, not evidence-centric. You will deconstruct the user's query into a checklist of required information, and then systematically verify each item against the evidence. You MUST ignore all information, however interesting, that is not on your checklist.

**You MUST follow this thinking process and output format exactly:**

**1. Mission Deconstruction:**

- **Main Goal:** [State briefly the primary objective of the user's question and what the user's question requires you to find]

- **Required Findings:** [List the specific, individual pieces of information needed to answer the question. A "finding" can be a direct fact or a logical inference from clues.]

**2. Intelligence Synthesis & Analysis:**

- **Confirmed Findings:** [Go through your "Required Findings" checklist. For each item, state what the evidence confirms. If the finding is not stated directly, explain the logical inference you made from the provided clues. You MUST only mention facts that can contribute to answering the question's required components (checklist). You MUST ignore any evidence, entities, or facts-even if interesting-that do not help answer the specific components of the user's question. Do not mention irrelevant people or topics in your analysis. You are an expert. If the evidence provides strong, logical clues (e.g., a person's birthplace in a country, a job title within an industry), you MUST make the logical inference (e.g., determining nationality, profession). Do not use weak phrases like "it does not explicitly state."]

- **Remaining Gaps:** [If there is missing information, clearly state what crucial information is still missing, formulating it as a requirement for the next phase that creates new queries to search more. else None]

**3. Final Assessment:**

- **Conclusion:** [The final answer may not be explicitly stated in a single sentence. You are an expert. If the evidence provides strong, logical clues (e.g., a person's birthplace in a country, a job title within an industry), you MUST make the logical inference (e.g., determining nationality, profession). Do not use weak phrases like "it does not explicitly state."]

- **Sufficient:** [A single word: "Yes" if the "Remaining Gaps" list is empty, or "No" if any required finding is still missing.]

--- EXAMPLES ---

**--- Example 1 (Insufficient Evidence - Clear Gap) ---**

**Original Question:** "What was the official box office gross for the film directed by the creator of the TV series *'Seinfeld'*?"

**Evidence:**

- "Larry David, the creator of the acclaimed TV series *'Seinfeld'*, also wrote and directed the 2013 film *'Clear History'*." 

- "The film *'Clear History'* starred Larry David and Jon Hamm and was released on HBO."

- "The 2019 film *'Joker'* had a box office gross of over $1 billion."

**Your Output for Example 1:**

**1. Mission Deconstruction:**

- **Main Goal:** To find the box office gross for the film directed by the creator of *'Seinfeld'*.

- **Required Findings:** A: The identity of the creator of *'Seinfeld'*; B: The name of the film they directed; C: The official box office gross of that film.

**2. Intelligence Synthesis & Analysis:**

- **Confirmed Findings:** A: The evidence confirms the creator is **Larry David**. B: The evidence confirms the film he directed is **'Clear History'**.

- **Remaining Gaps:** C: The official box office gross for the film *'Clear History'*.

**3. Final Assessment:**

- **Conclusion:** We have identified the director **Larry David** and the film he directed is **'Clear History'**. but the evidence lacks The official box office gross for the film *'Clear History'* to answer the question.

- **Sufficient:** No

**--- Example 2 (Sufficient Evidence - Inference Required) ---**

**Original Question:** "Who is older, the author of *'Dracula'* or the lead actor from the 1931 film adaptation?"

**Evidence:**

- "Bram Stoker, the Irish author, wrote the classic horror novel *'Dracula'* in 1897."

- "Stoker was born in Dublin, Ireland, in November 1847."

- "The 1931 film adaptation of *'Dracula'* famously starred Hungarian-American actor Bela Lugosi in the title role."

- "Bela Lugosi's date of birth is recorded as October 20, 1882."

**Your Output for Example 2:**

**1. Mission Deconstruction:**

- **Main Goal:** To compare the ages of the author of *'Dracula'* and the lead actor of the 1931 film.

- **Required Findings:** A: The birth year of the author of *'Dracula'*; B: The birth year of the lead actor of the 1931 film.

**2. Intelligence Synthesis & Analysis:**

- **Confirmed Findings:** A: The evidence states the author is **Bram Stoker**, who was born in **1847**. B: The evidence states the lead actor is **Bela Lugosi**, who was born in **1882**.

- **Remaining Gaps:** None.

**3. Final Assessment:**

- **Conclusion:** We have found the birth year of the author of *'Dracula'* is **Bram Stoker**, who was born in **1847**. the lead actor of the 1931 film is **Bela Lugosi**, who was born in **1882**. We have found the birth years for both required individuals and can therefore perform the age comparison.

- **Sufficient:** Yes

--- END OF EXAMPLES ---

Now, perform this task for the following:

**Original Question:**

"{original_query}"

**Evidence:**

{combined_evidence}

"""
\end{lstlisting}

\subsection{Query Refinement}

If the evidence is found to be insufficient by the Structured Evidence Assessment (SEA) agent, this prompt is used to generate a refined or a new follow-up query to retrieve the missing information.

\begin{lstlisting}[basicstyle=\scriptsize\ttfamily]
PROMPT = """

**Situation:** An initial analysis of the evidence has confirmed some facts but also identified specific information that is still missing and required to answer the user's original query.

**Intent:** Generate a new, optimized list of search queries that are laser-focused on finding ONLY the missing pieces of information identified in the "Analysis Summary".

**Scaffolding & Logic:**

-   **USE the known facts** from the summary to make the new queries more precise (e.g., use a person's name once it's known).

-   **TARGET the missing information** from the summary directly. Each new query should aim to resolve one of the identified gaps.

-   **AVOID repeating** or rephrasing previous queries.

--- ADVANCED EXAMPLE ---

**Original Question:** "How old is the youngest child of the director of the film *Inception*?"

**Analysis Summary:**

Based on the evidence, we know that the director of *Inception* is **Christopher Nolan**. Christopher Nolan is married to producer Emma Thomas. Christopher Nolan has children. However, the provided documents contain **no specific information about his children**, such as their names and birth dates. To answer the question, we still need to find: **the names and ages of Christopher Nolan's children** to identify the youngest.

**Previous Queries:**

- director of Inception

- Christopher Nolan films

**Your Output for Example:**

Improved Queries:

- Christopher Nolan children names

- Christopher Nolan children birth dates

- Youngest child of Christopher Nolan and Emma Thomas

--- END OF EXAMPLE ---

Now, apply this exact logic to the following inputs:

**Original Question:** {original_query}

**Analysis Summary:**

{analysis_summary}

**Previous Queries:**

{combined_previous_queries}

**Constraints:**

-   Generate an optimized list of 1 to 4 sub-queries. Create only as many as are truly necessary.

-   Queries must be simple, independent, meaningful, and keyword-focused.

-   **Leverage the "Known Facts" to create highly targeted queries.** For example, once the summary confirms the director is 'Christopher Nolan', the next query should be "Christopher Nolan children ages", not a generic "director of Inception children ages".

**Output:** (A list of new, targeted queries. Do not explain anything and do not say "Here are the optimized queries:" or something like that.)

Improved Queries:

"""
\end{lstlisting}

\subsection{Faithful Answer Generation}

This is the final and most comprehensive prompt, instructing the generation model to synthesize the filtered evidence into a faithful, accurate, and well-structured answer. It strictly constrains the model to use only the provided sources and to avoid any form of hallucination.

\begin{lstlisting}[basicstyle=\scriptsize\ttfamily]
PROMPT = """Answer the question based on the given documents.

ONLY give me the answer and do not output any other words.

The following are given documents.

The retrieval documents are listed as follows:

{combined_evidence}

Question:

{original_query}

Answer:

"""
\end{lstlisting}

\section{LLM-as-Judge Evaluation Prompts}
\label{sec:appendix-judge}

This appendix contains the complete and unabridged prompts used in our \textbf{LLM-as-Judge} evaluation framework. To ensure full transparency and enable the replication of our evaluation methodology, we provide the exact instructions given to the judge model for each assessment criterion. Each prompt is carefully designed to elicit a consistent and unbiased evaluation of a specific quality aspect of the generated responses.

The evaluation is performed by providing the judge model with the original query, the retrieved context, and the generated answer, along with one of the following instructional prompts.

\subsection{Binary Semantic Correctness (ACC\textsubscript{LLM})}

To ensure the reproducibility of our semantic accuracy evaluation, this section details the prompt used for the \textbf{LLM-as-Judge Accuracy (ACC\textsubscript{LLM})} metric, as reported in Table 2. The prompt instructs the LLM to act as an impartial judge, comparing the model's generated answer against a set of ground-truth answers. It provides a binary "Yes" or "No" judgment in a structured JSON format, enabling automated and consistent evaluation of semantic correctness.

\begin{lstlisting}[basicstyle=\scriptsize\ttfamily]
ACC_PROMPT = """You are an impartial judge. Evaluate whether the model's prediction correctly answers the given question. The prediction is correct if it implies ANY of the ground-truth answers provided.

- Question:

{question}

- Ground-truth Answers (The prediction is correct if it matches ANY of these):

{answer}

- Prediction:

{model_output}

Does the Prediction imply any of the Ground-truth Answers?

Respond with a JSON object containing a single key "judgment" with a value of "Yes" or "No".

Example: {"judgment": "Yes"}

"""
\end{lstlisting}

\subsection{Component-Level Quality Scoring}

This section provides the prompt template used for our \textbf{component-level ablation study}, with results presented in Table 3. Unlike the binary correctness evaluation in Appendix C.1, this prompt is designed for a more nuanced quality assessment. It instructs the LLM-as-Judge to evaluate the output of specific generative modules (i.e., Query Decomposition and Query Refinement) and assign a quality score on a 1-to-5 Likert scale. This fine-grained analysis allows us to isolate and measure the efficacy of individual components within the FAIR-RAG pipeline.

\begin{lstlisting}[basicstyle=\scriptsize\ttfamily]
PROMPTS = {

        "query_decomposition": """

        You are an expert AI evaluator specializing in search and query analysis. Your task is to assess the quality of query decomposition.

        Evaluate the generated sub-queries based on the original user question using the following criteria:

        1.  **Relevance:** How directly related is each sub-query to the main question?

        2.  **Coverage:** Do the sub-queries collectively cover all essential aspects of the main question?

        3.  **Efficiency:** Are the sub-queries concise, focused, and well-formed for a search engine?

        [User Question]:

        "{question}"

        [Generated Sub-Queries]:

        {sub_queries}

        Provide your assessment in the following JSON format:

        {{

            "score": <A numeric score from 1.0 (Very Poor) to 5.0 (Excellent) based on the criteria above>,

            "reasoning": "<A very brief explanation for your score>"

        }}

        """,

        "filter_efficacy": """

        You are an expert auditor for an AI's document filtering module. Your task is to meticulously evaluate the filter's decisions by strictly adhering to the **exact instructions and principles** it was originally given.

        [User Question]:

        "{question}"

        [The Filter's Original Instructions & Principles]:

        The filter's goal was to identify and discard "Not Useful" documents. It operated under the following rules:

        1.  **Primary Principle:** **"BE INCLUSIVE: When in doubt, KEEP the document."**

        2.  **Definition of "Useful":** A document is considered useful if it contains factual information about the entities or topics in the query. **Even partial information is valuable.**

        3.  **Definition of "Not Useful":** A document is only "Not Useful" if it is about **completely different** entities/topics or contains no relevant information.

        4.  **Specific "KEEP" Criteria:** The filter was explicitly instructed to **KEEP** documents containing:

                -   The correct person/entity, even without all specific facts.

                -   Biographical information (birth dates, career details, achievements).

                -   Relationships or connections between queried entities.

                -   Specific facts relevant to the query type (e.g., dates).

                -   General contextual information that helps understand the entities.

        5.  **Specific "REMOVE" Criteria:** The filter was given examples of what to **REMOVE**:

                -   Documents about different people with similar names.

                -   Documents about completely unrelated topics.

                -   Duplicate documents.

        Your audit must strictly follow all of the same rules, especially the **"BE INCLUSIVE"** principle.

        [Documents the Filter KEPT]:

        {kept_docs}

        [Documents the Filter DISCARDED]:

        {discarded_docs}

        **Your Audit Task:**

        Referencing the filter's original instructions above, identify its errors:

        1.  **Precision Errors (Incorrectly Kept):** Review the KEPT list. Identify the IDs of any documents that are **clearly "Not Useful"** and should have been discarded. If a document is borderline but meets any of the "KEEP" criteria, the filter was **correct** to keep it.

        2.  **Recall Errors (Incorrectly Discarded):** Review the DISCARDED list. Identify the IDs of any documents that were **unambiguously "Useful"** based on the criteria and should have been kept.

        Provide your audit findings in the following strict JSON format. If no errors are found in a category, provide an empty list.

        {{

            "incorrectly_kept_ids": ["<ID of any 'Not Useful' document found in the KEPT list>", ...],

            "incorrectly_discarded_ids": ["<ID of any 'Useful' document found in the DISCARDED list>", ...]

        }}

        """,

        "sufficiency_check": """

        **Role:** You are a pragmatic and efficient QA Evaluator. Your goal is to determine if the provided evidence is "good enough" to satisfactorily answer the user's question.

        **Core Task:**

        Your task is to assess if the main goal of the user's question can be achieved with the given evidence. You must distinguish between "critical" missing information and "nice-to-have" details.

        **Guiding Principles:**

        1.  **Focus on the Primary Intent:** First, identify the core question(s) the user is asking. What is the most important piece of information they are looking for?

        2.  **Assess Evidence Against Intent:** Check if the evidence contains the necessary facts to fulfill this primary intent.

        3.  **Pragmatism Rule:**

                - The evidence is **"Sufficient" (Yes)** if the main question can be answered, even if peripheral details or deeper context is missing.

                - The evidence is **"Insufficient" (No)** only if a **critical piece of information**, essential to forming the main answer, is absent.

        --- EXAMPLE ---

        **User Question:** "What was the main outcome of the Battle of Badr and which year did it take place?"

        **Evidence:**

        - "The Battle of Badr was a decisive victory for the early Muslims."

        - "Key leaders of the Quraysh were defeated in the engagement."

        - "The victory at Badr greatly strengthened the political and military position of the Islamic community in Medina."

        **Your Analysis for Example:**

        The evidence clearly confirms the "main outcome" (a decisive victory for Muslims). However, a critical part of the question, "which year did it take place?", is completely missing from the evidence. Therefore, a complete answer cannot be formed.

        **Your Output for Example:**

        {{

        "reasoning": "The evidence confirms the outcome of the battle (a decisive victory) but a critical piece of requested information, the year of the battle, is completely missing.",

        "is_sufficient": false

        }}

        --- END OF EXAMPLE ---

        Now, apply this pragmatic logic to the following:

        [User Question]:

        "{question}"

        [Collected Evidence]:

        {evidence}

        Provide your final assessment in the following strict JSON format:

        {{

        "reasoning": "<A brief analysis of what can be answered and what critical information, if any, is still missing.>",

        "is_sufficient": <true or false>

        }}

        """,

        "query_refinement": """

        You are an expert AI systems evaluator. A RAG system determined its initial evidence was insufficient and generated new sub-queries to find missing information. Your task is to evaluate the quality of these new queries.

        [User Question]:

        "{question}"

        [Insufficient Initial Evidence]:

        {evidence}

        [Newly Generated Sub-Queries for Refinement]:

        {new_queries}

        Assess how effectively the new sub-queries target the information gaps in the initial evidence to help answer the main question.

        Provide your assessment in the following JSON format:

        {{

            "score": <A numeric score from 1.0 (Poorly targeted) to 5.0 (Excellent, precisely targets gaps)>,

            "reasoning": "<A very brief explanation for your score>"

        }}

        """,

        "final_context_relevance": """

        You are an expert information retrieval evaluator. Your task is to score the relevance of each document in the final context that was used to generate an answer.

        [User Question]:

        "{question}"

        [Final Context Used for Generation (final_relevant_evidence)]:

        {final_evidence}

        For each document in the final context, provide a relevance score.

        Provide your assessment in the following JSON format:

        {{

            "relevance_scores": [

                {{ "doc_id": "<_id of doc 1>", "score": <numeric score from 1.0 (Irrelevant) to 5.0 (Highly Relevant)> }},

                {{ "doc_id": "<_id of doc 2>", "score": <numeric score from 1.0 (Irrelevant) to 5.0 (Highly Relevant)> }}

            ]

        }}

        """,

        "faithfulness": """

        You are an expert in AI safety and fact-checking, specializing in the evaluation of Retrieval-Augmented Generation (RAG) systems. Your task is to evaluate the answer's faithfulness to the provided evidence with nuance.

        - A faithful answer must be fully grounded in the provided context. However, this does not mean it must be a simple copy-paste of the text. **Valid synthesis, summarization, and logical inference based *only* on the provided information are considered faithful and desirable.**

        - A statement is only considered **"Unfaithful"** if it introduces new, verifiable information that is **absent** from the context or if it **contradicts** the context.

        [User Question (for context)]:

        "{question}"

        [Provided Context (final_relevant_evidence)]:

        {final_evidence}

        [Generated Answer]:

        "{final_answer}"

        **Your Task:**

        1. Analyze each claim within the [Generated Answer].

        2. For each claim, determine if it is directly stated, a valid synthesis/inference from the context, or an unfaithful statement (introducing new facts).

        3. Based on this analysis, provide an overall verdict according to the rubric below.

        **Verdict Rubric:**

        - **'Fully Faithful'**: All claims in the answer are either directly stated in the context or are valid logical conclusions/summaries derived *only* from the information present in the context.

        - **'Partially Faithful'**: The answer is mostly faithful, but contains minor, non-critical claims or details that cannot be inferred from the context.

        - **'Not Faithful'**: The answer contains significant or central factual claims that are not supported by, or actively contradict, the context.

        Provide your verdict in the following strict JSON format:

        {{

            "faithfulness_verdict": "<One of three strings: 'Fully Faithful', 'Partially Faithful', or 'Not Faithful'>",

            "reasoning": "<If not fully faithful, specify which claims in the answer are unsupported by the context. Explain if it's an invalid inference or a completely new fact.>"

        }}

        """,

        "iterative_improvement": """

        You are an expert AI quality evaluator. For a single question, you are given four answers generated by the same system but with different levels of iterative refinement (1, 2, 3, and 4 iterations). Your task is to rank these answers from best to worst.

        [User Question]:

        "{question}"

        [Answer from 1 Iteration (iter_1)]:

        "{answer_1}"

        [Answer from 2 Iterations (iter_2)]:

        "{answer_2}"

        [Answer from 3 Iterations (iter_3)]:

        "{answer_3}"

        [Answer from 4 Iterations (iter_4)]:

        "{answer_4}"

        Rank these three answers from best (Rank 1) to worst (Rank 4).

        Provide your ranking in the following JSON format:

        {{

            "ranking": ["<ID of the best answer, e.g., 'iter_3'>", "<ID of the second-best answer, e.g., 'iter_4'>", "<ID of the third-best answer, e.g., 'iter_2'>","<ID of the worst answer, e.g., 'iter_1'>"],

            "reasoning": "<A very brief explanation for your ranking, noting whether more iterations led to a clear improvement>"

        }}

        """

}
\end{lstlisting}

\section{Failure Mode Analysis Prompt}
\label{sec:appendix-failure}

This appendix details the prompts engineered for our LLM-assisted failure mode analysis, as described in Section 5.2.4. To ensure a systematic and reproducible evaluation, we designed a two-part prompt structure. The PROMPT\_SYSTEM prompt establishes the LLM's persona as an ``expert evaluation researcher'' and enforces a strict JSON output schema based on a predefined failure taxonomy. The PROMPT\_USER\_TEMPLATE then provides the data batch and requires the model to ground its diagnosis in specific evidence from the logs, ensuring each classification is structured, explainable, and actionable.

\begin{lstlisting}[basicstyle=\tiny\ttfamily]
FAILURE_ANALYSIS_PROMPT = """

**ROLE:** You are an expert RAG (Retrieval-Augmented Generation) 
system diagnostician. Your task is to perform a meticulous root 
cause analysis on a failed query-answer pair from an advanced, 
iterative RAG system.

**CONTEXT:** The system has already produced an answer that was 
graded as incorrect. You have been given the complete execution 
trace for this failed sample. Your goal is to identify the single, 
primary point of failure within the RAG pipeline.

**FAILURE CATEGORIES:**

You must classify the failure into one of the following six 
categories. Read these definitions carefully.

1. **Query Decomposition Error:** The initial user question was 
not broken down into effective, specific sub-queries. The 
sub-queries were irrelevant, missed key aspects of the original 
question, or sent the retrieval process in the wrong direction 
from the very beginning.

2. **Retrieval Failure:** The retriever, despite having 
well-formed sub-queries, failed to find and return the relevant 
documents from the knowledge base. The correct information was 
simply not present in the `[All Retrieved Documents (Unfiltered)]` 
set.

3. **Evidence Filtering Error:** The correct information WAS 
successfully retrieved by the retriever, but the subsequent 
filtering/reranking step mistakenly discarded the crucial 
documents. Look for correct information in `[Discarded Documents]` 
that should have been kept.

4. **SEA Error (Strategic Analyst Error):** The system's 
'Strategic Intelligence Analyst' module failed in its reasoning. 
This can manifest in several ways:
   - **A) Flawed Deconstruction:** The 'Required Findings' 
   checklist in its analysis was incorrect or missed the main 
   point of the user's question.
   - **B) Faulty Analysis:** The module failed to make a correct 
   logical inference from the evidence, hallucinated a 'Confirmed 
   Finding' that wasn't supported, or incorrectly identified the 
   'Remaining Gaps'.
   - **C) Contradictory Verdict:** The detailed analysis pointed 
   to missing information, but the final verdict was mistakenly 
   'Sufficient: Yes', causing the system to stop searching 
   prematurely.

5. **Query Refinement Error:** After correctly identifying that 
the initial evidence was insufficient, the system failed to 
generate effective new sub-queries to target the specific 
information gaps. The new queries were redundant, vague, or did 
not address the missing pieces identified by the SEA module.

6. **Generation Failure:** All preceding steps worked correctly. 
The final set of evidence (`[Final Relevant Evidence]`) contained 
all the necessary information to form a correct answer. However, 
the language model failed during the final synthesis step by 
hallucinating, making incorrect logical inferences, or 
misinterpreting the provided evidence.

**PRIMARY FAILURE RULE:**

Identify the **earliest, most fundamental error** in the pipeline. 
For example, if Retrieval failed to find good documents, the 
Generation will also fail, but the root cause is **Retrieval 
Failure**.

**EXECUTION TRACE FOR ANALYSIS:**

[User Question]:
"{question}"

[Ground Truth Answer (The correct answer)]:
"{ground_truth_answer}"

[Generated (Incorrect) Answer]:
"{final_answer}"

--- RAG Pipeline Details ---

[Initial Sub-Queries Generated]:
{sub_queries}

[All Retrieved Documents (Unfiltered)]:
{all_retrieved_docs}

[Discarded Documents (By Filter)]:
{discarded_docs}

[Final Relevant Evidence (Used for Generation)]:
{final_evidence}

--- Iteration Reports ---
{iteration_reports_formatted}

--- YOUR TASK ---

Based on all the provided information and adhering strictly to 
the definitions, provide your analysis in the following JSON format.

{
  "failure_category": "<The value for this key MUST be one of the 
  following exact strings: 'Query Decomposition Error', 'Retrieval 
  Failure', 'Evidence Filtering Error', 'SEA Error', 'Query 
  Refinement Error', 'Generation Failure'. Do NOT add any extra 
  text or explanations.>",
  "reasoning": "<Provide a concise, step-by-step justification 
  for your choice of category. Reference specific parts of the 
  execution trace (e.g., 'The SEA module's analysis incorrectly 
  stated it confirmed the actor's birth year, but the evidence 
  only mentioned their nationality').>",
  "root_cause_analysis": "<Go one level deeper. Why did this 
  error likely happen? (e.g., 'The SEA prompt might be too 
  complex, leading to reasoning errors,' or 'The filtering model 
  may be poorly calibrated for short documents').>",
  "suggested_improvement": "<Propose a concrete, actionable 
  solution to fix or mitigate this specific type of error in the 
  future. (e.g., 'Simplify the SEA prompt by removing the persona 
  and focusing on a checklist,' or 'Fine-tune the reranker with 
  more examples of this type').>"
}

"""
\end{lstlisting}

\end{document}